\documentclass[Afour,sageh,times]{sagej}

\usepackage{moreverb,url}
\usepackage{nameref}
\usepackage{color, colortbl}
\usepackage{xcolor}
\usepackage{csquotes}
\usepackage{bigstrut}
\usepackage{multirow}

\newcommand{\p}[1]{\smallskip \noindent \textbf{{#1}.}}
\newcommand{\eq}[1]{Equation~(\ref{eq:#1})}
\newcommand{\fig}[1]{Figure~\ref{fig:#1}}
\usepackage{balance}

\usepackage{hyperref}
 \hypersetup{
     colorlinks=true,
     linkcolor=orange,
     filecolor=orange,
     citecolor=orange,      
     urlcolor=orange,
     }
\usepackage{perpage} %the perpage package
\MakePerPage{footnote} %the perpage package command 
\usepackage{float}
\usepackage{wrapfig}

\begin{document}

%%%%%%%%%%%%%%%%%%%%%%%%%%%%%%%%%%%%%%%%%%%%%%%%%%%%%%%%%%%%%%%%%%%%%%%%%%%%%%%%

\runninghead{Keely et al.}

\title{Kiri-Spoon: A Kirigami Utensil for Robot-Assisted Feeding}

\author{Maya Keely\affilnum{1}, Brandon Franco\affilnum{1}, Casey Grothoff\affilnum{1}, Rajat Kumar Jenamani\affilnum{2}, Tapomayukh Bhattacharjee\affilnum{2}, Dylan P. Losey\affilnum{1}, and Heramb Nemlekar\affilnum{1}}

\affiliation{\affilnum{1}Virginia Tech, Department of Mechanical Engineering\\
\affilnum{2}Cornell University, Department of Computer Science}

\corrauth{Maya Keely, 
Department of Mechanical Engineering,
Virginia Tech, 
Blacksburg, VA,
24060, USA.}

\email{\texttt{mayakeely@vt.edu}}

%%%%%%%%%%%%%%%%%%%%%%%%%%%%%%%%%%%%%%%%%%%%%%%%%%%%%%%%%%%%%%%%%%%%%%%%%%%%%%%%
\begin{abstract}

For millions of adults with mobility limitations, eating meals is a daily challenge.
A variety of robotic systems have been developed to address this societal need.
These robots serve as a proxy for the human's arm: the user inputs the food they want to eat, and the robot autonomously picks up that food and brings it to the user's mouth.
Unfortunately, end-user adoption of robot-assisted feeding is limited, in part because existing devices are unable to seamlessly grasp, manipulate, and feed diverse foods.
Recent works seek to address this issue by creating new \textit{algorithms} for food acquisition and bite transfer.
In parallel to these algorithmic developments, however, we hypothesize that \textit{mechanical intelligence} will make it fundamentally easier for robot arms to feed humans.
We therefore propose Kiri-Spoon, a soft utensil specifically designed for robot-assisted feeding.
Kiri-Spoon consists of a spoon-shaped kirigami structure: when actuated, the kirigami sheet deforms into a bowl of increasing curvature.
Robot arms equipped with Kiri-Spoon can leverage the kirigami structure to wrap-around morsels during acquisition, contain those items as the robot moves, and then compliantly release the food into the user's mouth.
Overall, Kiri-Spoon combines the familiar and comfortable shape of a standard spoon with the increased capabilities of soft robotic grippers.
In what follows, we first apply a stakeholder-driven design process to ensure that Kiri-Spoon meets the needs of caregivers and users with physical disabilities.
We next characterize the dynamics of Kiri-Spoon, and derive a mechanics model to relate actuation force to the spoon's shape.
The paper concludes with three separate experiments that evaluate (a) the mechanical advantage provided by Kiri-Spoon, (b) the ways users with disabilities perceive our system, and (c) how the mechanical intelligence of Kiri-Spoon complements state-of-the-art algorithms.
Our results suggest that Kiri-Spoon advances robot-assisted feeding across diverse foods, multiple robotic platforms, and different manipulation algorithms.
Videos of our system and experiments are available here: \url{https://youtu.be/ZJpQREdTz80}

\end{abstract}


\maketitle

%%%%%%%%%%%%%%%%%%%%%%%%%%%%%%%%%%%%%%%%%%%%%%%%%%%%%%%%%%%%%%%%%%%%%%%%%%%%%%%%

\section{Introduction} \label{sec:intro}

\begin{figure*}[t]
    \begin{center}
        \includegraphics[width=2\columnwidth]{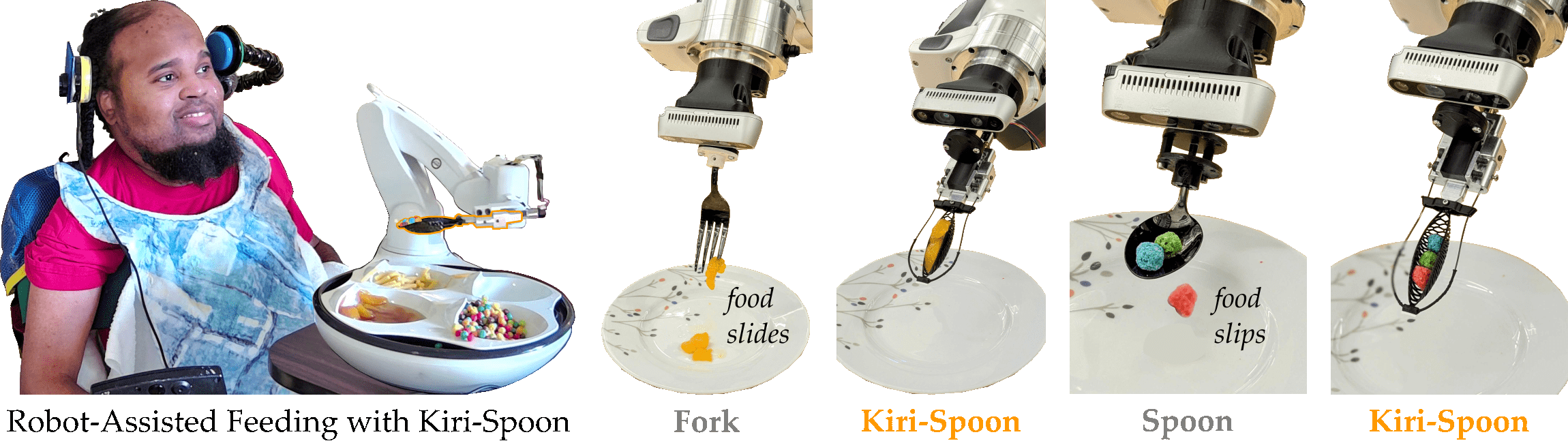}
        \caption{Kiri-Spoon is a spoon-shaped kirigami utensil specifically designed for robot-assisted feeding. (Left) Robot arms equipped with Kiri-Spoon can robustly acquire foods from the plate, safely carry those morsels to the human, and then seamlessly transfer items into the user's mouth. (Right) It is challenging for robot arms to dexterously manipulate traditional utensils such as forks and spoons. By comparison, Kiri-Spoon makes the robot's task fundamentally easier by flexibly wrapping around the desired foods. This capability enables Kiri-Spoon to function as a fork (pinching foods) or as a spoon (scooping foods).}
        \label{fig:front1}
    \end{center}
    \vspace{-1em}
\end{figure*}

\begin{figure}[t]
    \begin{center}
        \includegraphics[width=1\columnwidth]{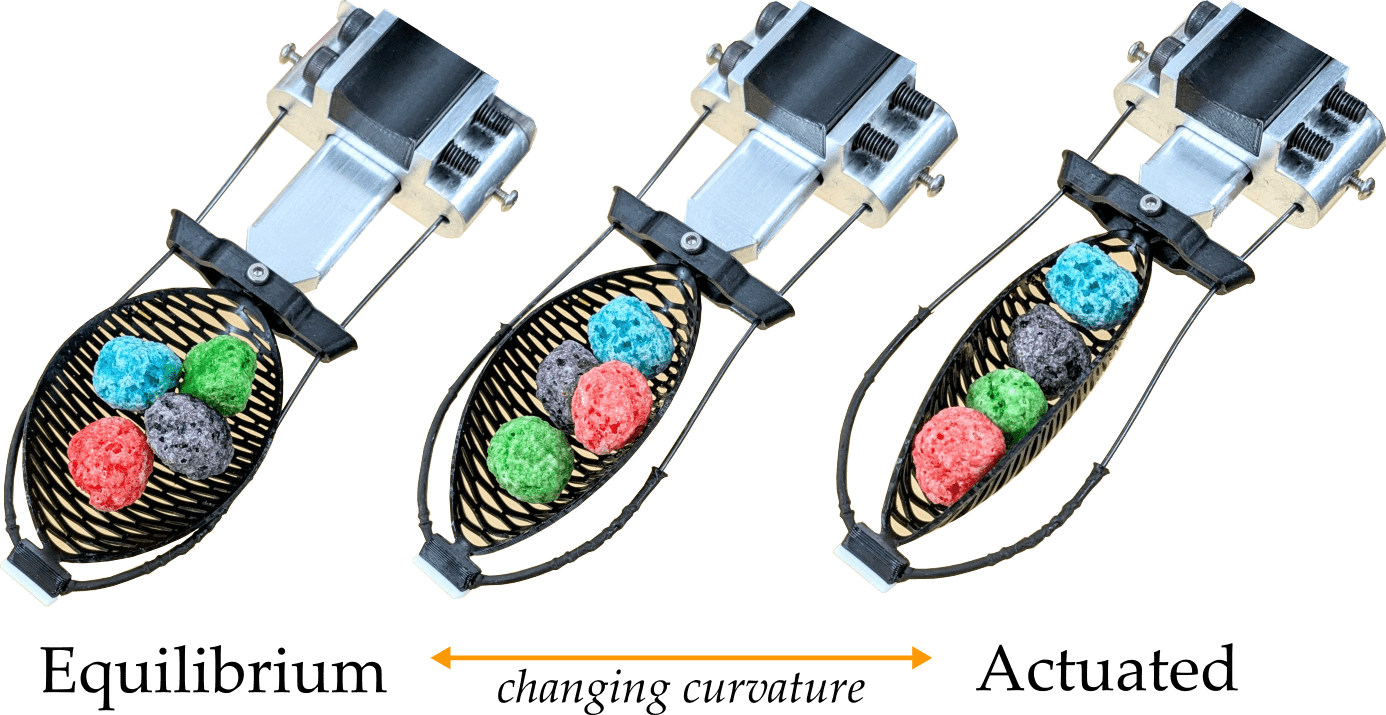}
        \caption{Actuating and releasing Kiri-Spoon. The core element of Kiri-Spoon is an elliptical kirigami sheet with discrete ribbons orthogonal to the applied forces. Retracting one end of Kiri-Spoon causes this $2$D sheet to buckle and form a $3$D bowl with increasing curvature, thereby encapsulating food items.}
        \label{fig:front2}
    \end{center}
    \vspace{-2em}
\end{figure}

Food is central to the human experience. 
Beyond its role in survival, health, and wellness, the ability to feed oneself fosters socialization, expresses identities, and marks significant moments \citep{nanavati2023design}. 
Unfortunately, there are almost $2$ million American adults living with disabilities who rely on a caregiver's assistance everyday in order to eat \citep{taylor2018americans}.
This reliance can lead to feelings of dependence among care recipients \citep{jacobsson2000eatingprocess, shune2020experience}, while also placing a significant workload on caregivers \citep{fleming2003caregiver, zenker2017exploring}. 
Assistive robots have the potential to help these users.
For example, wheelchair-mounted robot arms \citep{argall2018autonomy, kinova} and table-mounted feeding devices \citep{nanavati2023physically, obi} could enable operators to control their own meals and achieve modified independence.
Our work envisions robotic systems capable of picking up a bite of the user’s desired food and safely transferring it to their mouth.

To reach this future, existing work in robot-assisted feeding has focused on the algorithms the robot uses.
This includes learning and vision strategies for how the robot detects and skewers foods \citep{gordon2023towards, liu2024imrl, tai2023scone}, as well as control and motion planning for how the robot transfers that food item into the human's mouth \citep{ondras2023human, shaikewitz2023mouth, jenamani2024feel}.
All of these algorithmic features are necessary for a successful feeding system.
But the \textit{mechanical utensil} that the robot arm applies to pick up, carry, and deliver foods is also critical.
At first glance, it might seem obvious that robots should just leverage traditional utensils (e.g., forks and spoons).
People have been using spoons for at least $3,000$ years \citep{foote1934spoons}; these implements are optimized for humans, and prior works on assistive feeding have therefore equipped robots with familiar utensils.
But are rigid forks and spoons really the best utensil for a \textit{robot} to use?

Answering this question involves a balance between the robot's capability and the human's comfort. 
From the robot's perspective, utilizing traditional utensils like forks and spoons requires precise and careful manipulation.
To pick up a morsel the robot must dexterously coordinate the utensil's motion (e.g., skewering, scooping, twisting) in a way that is tailored to that specific food item.
Next --- after the food is grasped --- the robot needs to smoothly regulate its motion and orientation to prevent the food from spilling during transit (i.e., sliding off the fork or slipping out of the spoon).
New end-effectors mechanically address these challenges.
Recent soft grippers such as \cite{shintake2018soft, gafer2020quad, keely2024combining, li2019vacuum, glick2018soft} enhance robot capabilities by robustly encapsulating, holding, or adhering to diverse food items in ways that traditional forks or spoons cannot achieve.
But from the human's perspective, this increase in robot capability comes at the cost of convenience.
Today's soft end-effectors are not utensils: users cannot easily or comfortably transfer foods from these grippers into their mouths.
Instead, traditional utensils like forks or spoons are best suited for the human's needs --- people seamlessly take bites of foods from these familiar implements.

In this paper we propose to complement recent algorithmic advances in assistive robot arms by introducing a physical utensil specifically for robot-assisted feeding.
To augment the robot's capabilities while accounting for the human's perspective, our hypothesis is that:
\begin{center}\vspace{-0.3em}
\textit{Utensils for assistive robots should combine \\the enhanced \emph{functionality} of soft grippers with \\the comfortable \emph{form factor} of traditional utensils.}
\vspace{-0.3em}
\end{center}
Based on this hypothesis we introduce \textbf{Kiri-Spoon}: a \textbf{spoon}-like utensil featuring a soft \textbf{kiri}gami base (see \fig{front1}).
In its equilibrium state the Kiri-Spoon maintains the same size and shape as a traditional spoon, allowing users to take natural bites of food.
But when actuated, the Kiri-Spoon rapidly increases curvature, creating a bowl that robustly holds foods within its compliant manifold (see \fig{front2}).
Across bite acquisition, carrying, and transfer, Kiri-Spoon offers a \textit{mechanical advantage} to the process of robot-assisted feeding.
During acquisition, robot arms no longer need to make fine-grained skewering, scooping, or twisting motions: once the Kiri-Spoon is in contact with the desired morsel, we can simply actuate the utensil to grasp that item.
Similarly, when carrying the acquired foods, the robot does not have to maintain a steady speed or specific orientation: the actuated Kiri-Spoon encloses items so that they cannot easily slide or fall during transit.
Finally --- after bringing the food to the human --- the system reverts to a typical spoon form factor for human-friendly bite transfer.

Overall, this paper introduces, characterizes, and evaluates the first utensil specifically designed for robot-assisted feeding.
We make the following contributions\footnote{Note that a preliminary version of this work was published at the IEEE/RSJ Int. Conf. on Intelligent Robots and Systems \citep{keely2024kiri}. This paper is significantly different because here we redesign Kiri-Spoon alongside stakeholders, derive a detailed mechanics model, and then conduct multiple new experiments. These new experiments include autonomously acquiring diverse foods, evaluations with stakeholders, as well as comparing the effects of algorithmic and mechanical intelligence.}:

\p{Designing Kiri-Spoon}
We start by introducing Kiri-Spoon.
Our design process follows an iterative approach led by stakeholders --- including both caregivers and end-users.
Through interviews with occupational therapists and trials with people that have physical disabilities, we mutually arrive at Kiri-Spoon.
At its heart, the key component of Kiri-Spoon is a soft, elliptical sheet of plastic with parallel cuts (i.e., a $2$D kirigami sheet).
When this kirigami structure is pulled on both ends it deforms into a $3$D bowl with increasing curvature; when released, it returns to a flat spoon shape.

\p{Modeling the Mechanics and Geometry}
We next derive physics models to capture the relationship between applied forces and kirigami geometry.
To reach these models we separately analyze each interdependent component of the kirigami system: i) the boundary ribbon that surrounds the edge of our spoon, ii) the discrete ribbons that form the center of our spoon, and iii) the mesh ribbons that create our connected base.
We then combine these terms to predict the amount of tensile force needed to cause the Kiri-Spoon to wrap around food items.
Our validation tests show that the resulting model is an accurate lower bound across Kiri-Spoon designs with varying materials, thickness, and size.

\p{Equipping Assistive Robots with Kiri-Spoon}
Our work is motivated in part by the need to develop a utensil for assistive robot arms.
Accordingly, we next quantify how effectively robots can leverage our proposed utensil as compared to traditional forks and spoons.
We perform these experiments in ideal conditions where the robot arm uses each utensil to autonomously pick up a diverse set of known foods (e.g., carrots, tofu, soup, and cereal).
Importantly, we show that robots can employ Kiri-Spoons both as a fork (pinching carrots or tofu) and as a spoon (scooping soup or cereal).
We also identify the types of foods that are the failure cases for our Kiri-Spoon --- large, flat morsels such as bread or lettuce.

\p{Evaluating across Users with Disabilities}
Our work is also motivated by the need to make a utensil that stakeholders actually want to use.
To explore how stakeholders perceive our final system, we attached Kiri-Spoon to a commercially available feeding device \citep{obi}.
We then brought this assistive device to a local center for adults living with physical disabilities.
A caregiver and four participants tested the system with a standard spoon and with our proposed Kiri-Spoon.
Overall, the participants subjectively rated Kiri-Spoon to be about as comfortable as a traditional spoon.
In addition, their objective results and subjective responses show that Kiri-Spoon led to a higher success rate: foods were transferred from the table to the human's mouth more frequently with the Kiri-Spoon.

\p{Combining Mechanical and Algorithmic Advances}
We conclude this paper by exploring how both mechanical and algorithmic components contribute to the effectiveness of robot-assisted feeding.
We conduct a study along two axes: (a) the utensil the robot manipulates, and (b) the algorithm the robot uses to manipulate that utensil.
Sixteen users without disabilities operate an assistive robot to grasp, carry, and then eat multiple foods.
At one extreme, we equip the robot with standard forks and spoons, and the human manually teleoperates the robot throughout the entire eating process.
At the other extreme, the robot leverages Kiri-Spoon and state-of-the-art assistive algorithms to detect, acquire, and then carry the desired foods to the user.
We demonstrate that the combination of Kiri-Spoon and recent algorithmic advances leads to a more effective system than alternatives which improve only the hardware or only the software.

\section{Related Work} \label{sec:related}

\subsection{History of Robot-Assisted Feeding}

For people living with physical disabilities who currently rely on caregivers for meals, assistive technology offers an empowering tool towards increased independence \citep{argall2018autonomy,nanavati2023physically,van2024impact}.
Accordingly, robot-assisted feeding has a rich history.
In general, assistive feeding devices autonomously pick up morels of food, carry that food to the human, and then hold the food in place while the user takes a bite.
The human operator typically has control over which types of foods --- and how much of those foods --- the robot feeds to them.
The first systems were introduced in the 1970s and 1980s, including the Morewood Spoon Lifter \citep{vapc1977}, the Robotic Arm/Worktable System for self-feeding \citep{seamone1985early}, and Handy-1 \citep{topping2000overview}. 
These devices were extensively evaluated through user studies involving individuals with mobility limitations \citep{phillips1987feasibility}, paving the way for commercial feeding robots in the 1990s and early 2000s. 
Notable examples include the Winsford Feeder, My Spoon, Neater Eater, Bestic Arm, Meal Buddy, and Obi \citep{isira2015survey}.

The resulting devices leverage traditional utensils --- i.e., forks and spoons --- to acquire, hold, and then transfer morsels to the human.
In practice, however, they often either (a) struggle to pick up foods from plates and bowls, or (b) unintentionally spill food while carrying it to the user.
This fundamentally limits the effectiveness and convenience of today's assistive eating technology.
Overall, the lack of proficiency, comfort, and adaptability has contributed to limited adoption of robot-assisted feeding: out of all the devices mentioned above, only Neater Eater and Obi \citep{obi} remain commercially available (but not widely used).

\begin{figure*}[t]
    \begin{center}
        \includegraphics[width=2\columnwidth]{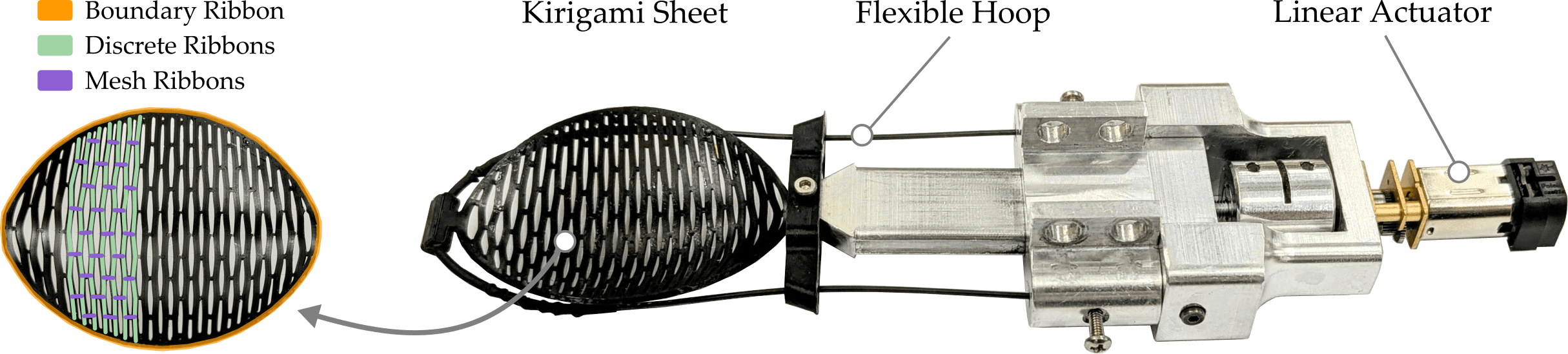}
        \caption{Design of Kiri-Spoon. (Left) A kirigami sheet is used to grasp, hold, and release food items. This sheet is composed of multiple ribbons: a boundary ribbon that surrounds the sheet, discrete ribbons that form the base of the spoon, and mesh ribbons that interconnect the discrete ribbons. (Right) The kirigami sheet is supported on one end by a flexible hoop. The other end is extended or retracted by a 1-DoF linear actuator. During eating, users interact with the flexible hoop and kirigami sheet.}
        \label{fig:design1}
    \end{center}
    \vspace{-1em}
\end{figure*}

\subsection{Autonomous Algorithms for Feeding}

To address these fundamental shortcomings, recent research has focused on developing autonomous algorithms for food acquisition and transfer. 
For example, approaches such as \cite{feng2019robot, gordon2023towards, sundaresan2023learning2, liu2024imrl, tai2023scone} study how a robot arm should manipulate a single utensil (e.g., a fork or spoon) to effectively pick up food items from a plate. 
These works highlight that humans typically employ a few basic motions during food acquisition with traditional Western utensils: stabbing, scooping, or twirling. 
Robots adapt these motion patterns to specific food items by reasoning over real-time visual and haptic data.
In scenarios where acquiring the desired food involves multiple steps --- such as pushing meatballs aside in order to access pasta noodles underneath --- learning-based approaches have been proposed to tackle complex, long-horizon manipulation tasks \citep{grannen2022learning, jenamani2024flair, bhaskar2024lava, sundaresan2023learning1, ha2024repeat}.

Once the robot acquires the desired food item, it must safely and comfortably transfer that food to the human’s mouth. 
Research on bite transfer seeks to coordinate the robot's utensil throughout this process. 
This includes deciding when to feed the user \citep{ondras2023human}, bringing the food item to the user's mouth so they can lean forward to take a bite \citep{fang2018rgb, ricardez2018quantitative, gallenberger2019transfer, belkhale2022balancing}, or placing food items inside the user's mouth if necessary \citep{park2017multimodal,shaikewitz2023mouth,jenamani2024feel}. 
Existing works leverage visual data from the robot's in-hand camera to perceive the user’s mouth, motion planning to control the arm, and haptic sensing to transfer food without applying unsafe forces. 
Some systems also rely on the user to partially guide the robot arm; for example, operators can leverage natural language or teleoperation inputs to correct the robot's motion throughout food acquisition and bite transfer \citep{padmanabha2024voicepilot, rea2022still, canal2016personalization, losey2022learning, jonnavittula2024sari}.

In summary, state-of-the-art research focuses on algorithms for autonomously picking up and transferring food items. 
But within these works the robot uses traditional forks and spoons --- and this inherently makes it more complicated for the robot to acquire and transfer foods.
Instead of forcing robot arms to dexterously manipulate rigid utensils that were originally designed for humans, we will focus on creating a utensil specifically for robot-assisted feeding.

\subsection{Soft Grippers for Food Manipulation}

Looking outside the domain of assistive eating, there are a variety of robotic grippers already built for food processing and handling
\citep{wang2022challenges}.
These grippers enhance the robot's ability to pick up and hold different food items by introducing new grasping mechanisms.
For example, today's robots can leverage shape-changing structures to enclose morsels within a soft end-effector \citep{shintake2018soft, brown2010universal, gafer2020quad, wang2017prestressed}.
Alternatively, other grippers utilize adhesive materials to cause foods to stick to the robot's surface \citep{keely2024combining, glick2018soft, hu2021soft}, or create vacuums that maintain a suction pressure on the desired morsel \cite{li2019vacuum, hao2020multimodal}.
Each of these grasping mechanisms provides advantages during acquisition: robots equipped with soft grippers can pick up, hold, and manipulate diverse food items more effectively than when using traditional rigid utensils.

However, state-of-the-art soft grippers are not designed for assistive eating applications, making it challenging to transfer food from the robot’s gripper to the user’s mouth. 
For instance, the food handling mechanisms proposed by \cite{gafer2020quad}, \cite{keely2024combining}, and \cite{wang2017prestressed} would require the robot to drop morsels directly onto the user, which is impractical for feeding. 
Overall, food-handling grippers differ too greatly from traditional utensils for comfortable human use.
At the other extreme, some works introduce minor modifications to traditional utensils. 
For example, \cite{sundaresan2023learning2}, \cite{shaikewitz2023mouth}, and \cite{jenamani2024flair} add active degrees of freedom to a fork, while \cite{eli} incorporates passive degrees-of-freedom into a spoon for improved balance. 
Although these modified utensils are more user-friendly --- allowing the user to take a bite directly --- they lack the advanced capabilities of food-handling grippers, limiting their functionality and leading to the same challenges faced with traditional utensils (e.g., food slipping from the utensil).

Kiri-Spoon lies between these two mechanical extremes: it bridges the gap between the functionality of soft, shape-changing grippers and the familiar form of a traditional spoon. 
Our design achieves this balance by leveraging kirigami structures. 
While kirigami has been incorporated into previous grippers \citep{buzzatto2024multi, yang2021grasping}, it has not yet been utilized for robot-assisted feeding. 
We aim to advance the state-of-the-art by designing, characterizing, and testing a human-friendly utensil tailored for robot-assisted feeding applications.

\section{Kiri-Spoon Design} \label{sec:design}

In this section we present our design for Kiri-Spoon\footnote{The design files for Kiri-Spoon are available at: \url{https://github.com/VT-Collab/Kiri-Spoon}}.
Here we apply our fundamental hypothesis: utensils for assistive feeding should be similar to traditional utensils in form, and similar to soft grippers in functionality.
By leveraging this hypothesis we ultimately reach a novel utensil with a spoon's form factor that utilizes an actuated kirigami structure to encapsulate and release food items (see \fig{design1}).
To ensure that this design is comfortable and meets the needs of users, we collaborate with stakeholders.
To ensure that this design is food-safe and effective at grasping diverse morsels, we engineer a morphing bowl with controllable curvature.
In what follows we describe our specific problem setting (Section~\ref{sec:K1}), how stakeholders were involved in the iterative design process (Section~\ref{sec:K2}), and the individual components of our resulting Kiri-Spoon (Section~\ref{sec:K3}).

\subsection{Problem Statement} \label{sec:K1}

We consider settings where a human with mobility limitations is leveraging an assistive robot arm to eat their everyday meal.
This meal is placed on a table in front of the user.
Next to the user --- either attached to their wheelchair or mounted on the table --- is an assistive robot arm.
This robot is equipped with cameras to perceive the environment, as well as a utensil to manipulate food items.
Consistent with standard practices for assistive eating, we assume that the foods are already in bite-sized morsels: i.e., the robot does not need to cut any items \citep{bhattacharjee2020more}.
At each iteration the human specifies which food item they want to eat, and the robot attempts to pick up that morsel, carry it to the human, and then help transfer it into the human's mouth.
From the robot's perspective, the system should successfully grasp, hold, and deliver the desired foods without spilling or dropping them.
From the human's perspective, the system should be safe, comfortable, and intuitive.

\subsection{Stakeholder-driven Iterative Design} \label{sec:K2}

In order for robot-assisted feeding to be successful, it is critical that these systems are designed with and by the stakeholders.
Prior works have therefore integrated caregivers and adults with disabilities into the design process \citep{nanavati2023design,ljungblad2023applying,pascher2021recommendations,wald2024mistakes,bhattacharjee2020more}.
Here we similarly adopt a stakeholder-centric approach.
Specifically, we interacted with occupational therapists and residents at The Virginia Home, a community for adults living with physical disabilities \citep{virginiahome}.

After collecting informal interviews about the types of assistive utensils that would be helpful and practical, we first developed an early iteration of Kiri-Spoon \citep{keely2024kiri}.
We then brought this iteration to The Virginia Home: here a set of $N=4$ users compared our initial Kiri-Spoon design to a standard plastic spoon.
Both spoons were held and manipulated by an Obi robot --- see Section~\ref{sec:userstudy} for more details on the experimental setup and protocol.
Overall, we leveraged Likert scale surveys and forced-choice questions to assess the user's perception of the utensil.
We also recorded objective measures for the amount of food transferred to the human's mouth and the number of times food items were unintentionally dropped.
Through this back-and-forth process we identified design steps needed for improved efficiency and comfort.
For example, our initial version of the Kiri-Spoon incorporated a rigid metal hoop to support the kirigami sheet.
Stakeholders indicated that this rigid hoop was uncomfortable when eating, and recommended a more compliant structure so that they could manipulate the utensil and foods within their mouth.
More generally, each of the key elements of our device --- including its geometry, materials, and functionality --- were reviewed or suggested by multiple stakeholders.
We therefore present the Kiri-Spoon as a stakeholder-led system, created in collaboration with members of our target population.

\begin{figure}[t]
    \begin{center}
        \includegraphics[width=0.8\columnwidth]{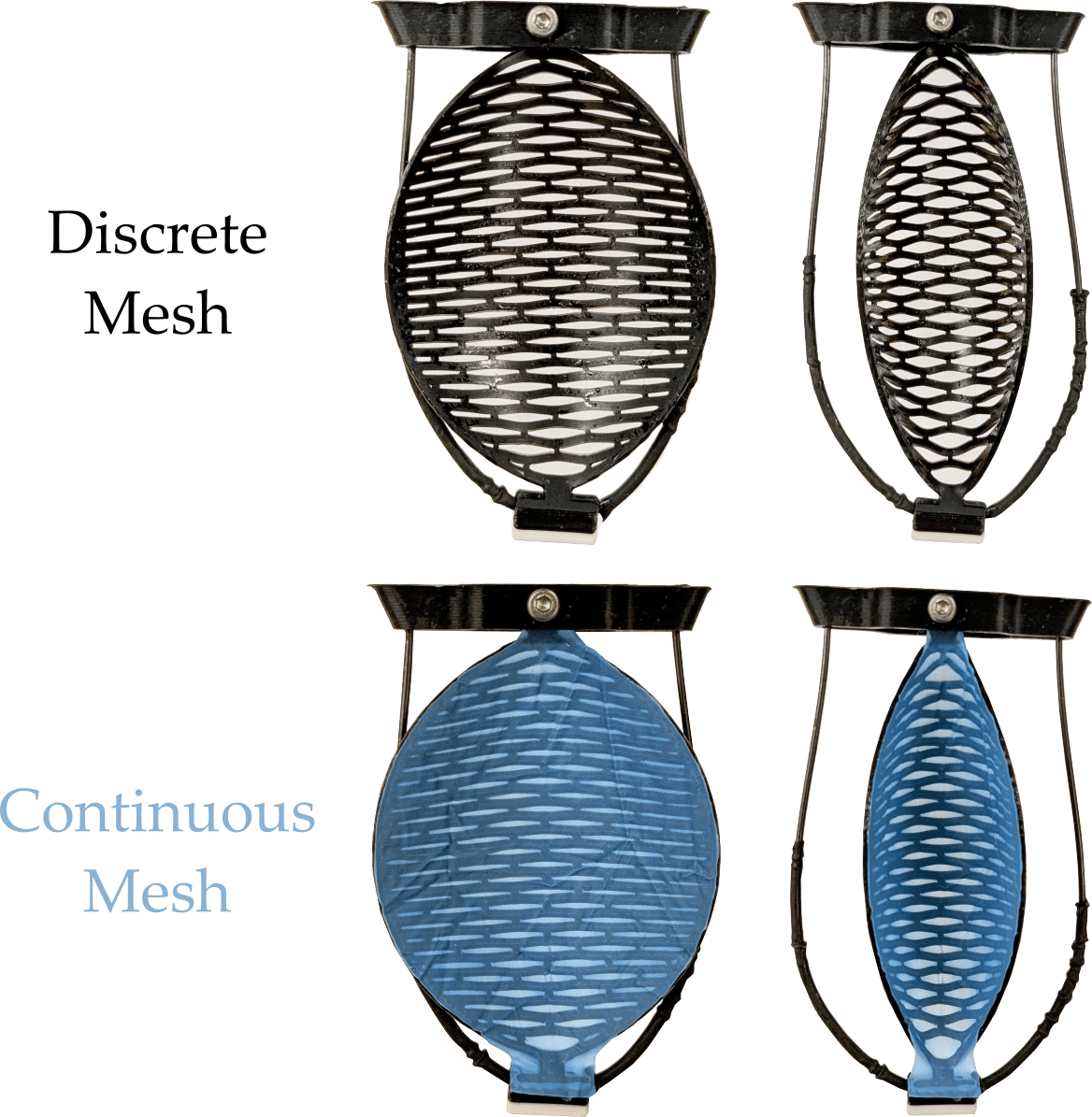}
        \caption{Two variations of Kiri-Spoon's mesh. (Top) For most foods a discrete mesh is sufficient. (Bottom) However, for liquid foods such as soups, a thin membrane can be mounted to the kirigami sheet. The resulting continuous mesh prevents liquids from falling out of the bottom of Kiri-Spoon.}
        \label{fig:design2}
    \end{center}
    \vspace{-1em}
\end{figure}

\begin{figure}[t]
    \begin{center}
        \includegraphics[width=1\columnwidth]{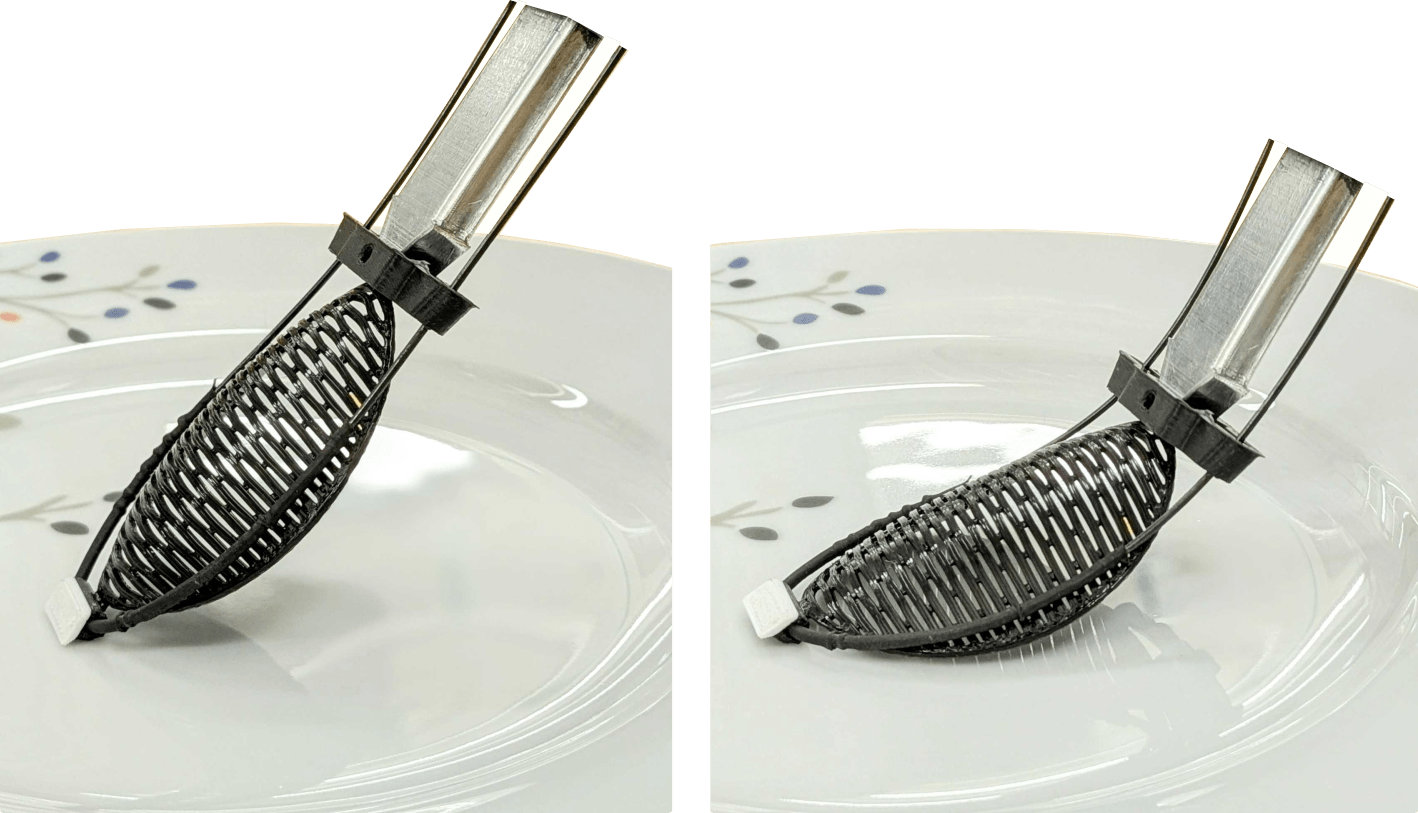}
        \caption{Demonstration of the flexible hoop. This flexibility is not only comfortable for users, but it also enables Kiri-Spoon to bend along the surface of plates and bowls. We leverage this flexibility to deploy Kiri-Spoon like a fork and pinch foods that are directly beneath the kirigami structure.}
        \label{fig:design3}
    \end{center}
    \vspace{-1em}
\end{figure}

\begin{figure*}[t]
    \begin{center}
        \includegraphics[width=1.0\textwidth]{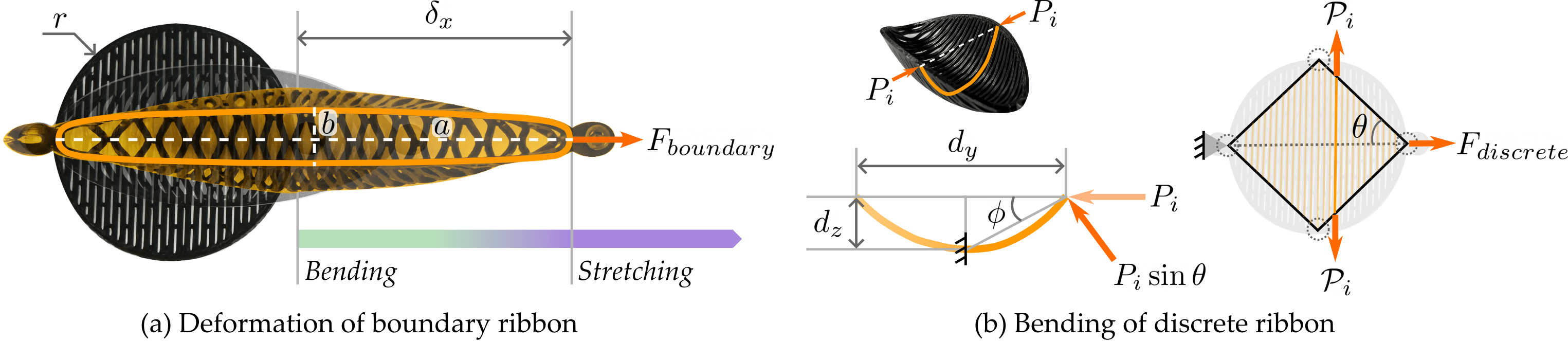}
        \vspace{-3 ex}
        \caption{Mechanics of the boundary and discrete ribbons under tensile load. 
        (a) $F_{boundary}$ is the tensile force component needed to bend the boundary ribbon. $\delta_{x}$ is the total displacement from its undeformed position. The boundary starts as a circle of radius $r$ and bends into an ellipse with semi-major axis $a$ and semi-minor axis $b$. As the ellipse becomes flat, the boundary begins to stretch.
        (b) The boundary applies a compressive force $P$ on the discrete ribbons, bending them into an arch. $\phi$ is the angle between $P$ and the bent discrete ribbons. In response to the boundary compression, each ribbon exerts an equal opposing force on the boundary. $F_{discrete}$ is the tensile force component needed to overcome this opposing force. 
        }
        \label{fig:dynamics}
        \vspace{-2 ex}
    \end{center}
\end{figure*}

\subsection{Kiri-Spoon Components} \label{sec:K3}

Our resulting design for Kiri-Spoon is outlined in \fig{design1}. 
This design consists of three essential elements: a \textbf{kirigami sheet} that forms the spoon, a \textbf{flexible hoop} that holds one end of the kirigami sheet, and a \textbf{linear actuator} that displaces the other end of the kirigami sheet.
Below we separately discuss each Kiri-Spoon component.

\p{Kirigami Sheet} The main component of Kiri-Spoon is a shape-morphing kirigami sheet.
This sheet is composed of multiple discrete ribbons (which form the center of the spoon), a boundary ribbon (which acts as the edge of the spoon), and a pattern of mesh ribbons (which create an interwoven base).
In its low-energy state our kirigami sheet is a flat, 2D ellipse with discrete ribbons parallel to the minor axis.
When the boundary ribbon is pulled orthogonally to these discrete ribbons, the discrete ribbons \textit{buckle} and the 2D sheet morphs into a 3D bowl.
The resulting structure resembles a spoon, but with two important distinctions.
First, the kirigami sheet is soft and deformable.
Second, by pulling or releasing the boundary layer we can control the curvature of our kirigami structure along a continuous spectrum.

In practice, the resulting kirigami sheet undergoes significant shape changes.
To achieve these changes we must manufacture the kirigami out of isotropic materials that deform uniformly when pulled.
Similarly, the materials should be ductile enough to extend when actuated, while also elastic enough to return to their original shape when released.
We therefore fabricate our kirigami structures with thermoplastic polyurethane (TPU), an inexpensive and \textit{food-safe} plastic that exhibits the desired material properties. 
The sheets are 3D printed to customize the geometry of the boundary ribbon, discrete ribbons, and mesh ribbons.
We note that these mesh ribbons can be discrete (as shown in \fig{design2}, Top) or one continuous structure (as shown in \fig{design2}, Bottom).
Using a continuous structure prevents liquid foods from falling out of the base of Kiri-Spoon.

\p{Flexible Hoop} The distal end of the kirigami sheet is mounted to a supporting hoop.
When the human eats from the Kiri-Spoon, this hoop often enters their mouth; and when Kiri-Spoon is actuated, this hoop applies axial forces to extend the kirigami sheet.
Correctly designing this hoop is therefore challenging, because the hoop must be a) flexible for the user's comfort, and also b) rigid enough to hold one end of the kirigami sheet in place.
In collaboration with our stakeholders, we found an effective trade-off between these goals by leveraging nickel titanium (i.e., nitinol) wire.
In our final design the flexible hoop is formed from a $1$mm diameter nitinol wire coated by a soft, food-safe plastic wrapper.
As shown in \fig{design3}, the resulting hoop bends when pressed against plates, bowls, or the human's mouth, and then reverts to its original shape after the contact ends.
This compliance is particularly useful when Kiri-Spoon is deployed as a fork: if pressed against a plate, the hoop bends so that the kirigami sheet is parallel to the desired food morsel.

\p{Linear Actuator} The proximal end of the kirigami sheet is attached to the output of a 1-DoF linear actuator.
This actuator applies controlled forces to extend or retract the boundary ribbon.
We construct the actuator out of a rotatory $12$V motor with an integrated encoder, and then connect that motor to a linear screw.
Rotating the motor in one direction applies tension to the kirigami sheet, and rotating the motor in the opposite direction releases this tension.
The actuator housing contains ball bearings for smooth, constrained motion; the system can fully extend or retract the kirigami sheet in under $2$ s.

\p{Integration} By combining the kirigami sheet, flexible hoop, and linear actuator, we reach our Kiri-Spoon design.
Kiri-Spoon can autonomously adjust the curvature of its spoon-like bowl to mechanically enclose or release food items.
The specific kirigami sheet actuated by Kiri-Spoon is modular; caregivers can replace this sheet with new 3D printed designs for different users or for different sets of foods.
For practical purposes, we emphasize that the flexible hoop and kirigami sheet are both \textit{washable} and \textit{food-safe}.
The total weight of the Kiri-Spoon used in our tests is $85$~g, and its volume at $15$~mm displacement is $\approx 8$~mL.
The life-cycle of the kirigami sheet depends on its materials and geometry: for example, in all our experiments we have not had TPU sheet A break.
However, we did notice some minor effects of repeated use.
After roughly $30$ min of repeated actuation, the semi-major axis of the kirigami sheet extends less than $3$ mm.
Even after long-term usage we have not experienced a permanent displacement of more than $6$ mm.

Once manufactured, Kiri-Spoon is then mounted at the end-effector using custom 3D printed attachments included in our online repository. 
In the following experiments we connect Kiri-Spoon to robot arms --- e.g., FrankaEmika \citep{frankaemika}, UR5 \citep{ur5} --- and an assistive feeding device --- e.g., Obi \citep{obi}.
The algorithm used to control that system must account for Kiri-Spoon in two ways: the timing when Kiri-Spoon is opened or closed, and the amount of curvature of the kirigami structure.
Similar to the geometry of the sheet, these control parameters can be adjusted for the specific user or food item.
In Section~\ref{sec:model} we will present a mechanics model of Kiri-Spoon.
Then, in Sections~\ref{sec:auto_experiment}--\ref{sec:finalstudy}, we will integrate the mechanical system with robot arms and assistive feedings algorithms.
\section{Mechanics Model} \label{sec:model}

In this section we develop a theoretical model of the kirigami sheet that forms the basis of Kiri-Spoon.
During feeding, this kirigami sheet is pulled perpendicular to the discrete ribbons in order to create a spoon-shaped structure (see \fig{design1} and \fig{design2}).
Intuitively, the force required to deform the kirigami sheet depends on its geometry and material properties. 
Here we seek to better understand this intuition --- specifically, we quantify the amount of tensile force needed to actuate Kiri-Spoon.

As shown in \fig{design1}, our kirigami structure is composed of three different types of interconnected ribbons that buckle and deform.
Hence, to reach an overall model, we must consider each component.
We individually analyze: (i) the force required to deform the boundary ribbon ($F_{boundary}$), (ii) the force needed to bend the discrete ribbons ($F_{discrete}$), and (iii) the force due to resistance from the mesh ribbons ($F_{mesh}$).
By combining each of these components, we ultimately reach a lower bound on the 
force applied to the kirigami sheet as a function of its shape.
% shape of the kirigami sheet as a function of the applied force.
Designers can leverage this mechanics model to select the materials, geometry, and actuator for their own Kiri-Spoons.

\subsection{Boundary Ribbon Deformation}\label{sec:boundary}

We start with the outer boundary ribbon that surrounds the kirigami sheet.
In its equilibrium state this boundary ribbon is a circle with radius $r$.
When tensile force is applied to the boundary, it deforms into an ellipse where its major axis is aligned with the direction of the applied force.
We show this process in \fig{dynamics}(a) --- let $a$ and $b$ denote the lengths of the semi-major and semi-minor axes, respectively. 

\p{Geometry} We first calculate the dimensions of the boundary ribbon.
These dimensions are practically important because food enters and exits the kirigami structure through this boundary ribbon --- and we can even utilize the boundary ribbon to pinch morsels.
For a displacement of $\delta_{x}$ in the direction of the tensile force, the length of the semi-major axis $a$ becomes $a = r + (\delta_{x}/2)$. 
To compute the semi-minor axis $b$, we assume that the length of the boundary ribbon (i.e., the perimeter) remains roughly constant during deformation.
This enables us to leverage a standard formula for the dimensions of an ellipse:
\begin{equation}
    \pi\left(3(a+b) - \sqrt{(3a+b)(a+3b)}\right) = 2\pi r \label{eq:perimeter}
\end{equation}
The left side of Equation~\ref{eq:perimeter} is the perimeter of the deformed ellipse, while the right side is the circumference of the initially circular boundary.
Solving this formula for $b$ completes the geometry of the boundary ribbon.

\p{Force} Given the geometry, we next compute the tensile force needed to deform the boundary ribbon.
We denote this overall force as $F_{boundary}$.
Intuitively, $F_{boundary}$ is what the actuator must apply in order to extend the semi-major axis in \fig{dynamics}(a).
As we will show, there are two components of this force: bending (at low displacement) and stretching (at high displacement).
Let $F_{bend}$ denote the bending component. 
Applying bending theory for circular rings \citep{timoshenko2012theory}, we reach:
\begin{equation}
    F_{bend} = \frac{4EI \delta_{x}}{r^{3}} \left(\pi - \frac{8}{\pi}\right)^{-1} \label{eq:f_bend}
\end{equation}
where $E$ is the Young's modulus and $I$ is the second moment of area of the boundary cross-section. 
We set the initial radius of curvature for the ring to be $r$, since $r$ is the radius of the initially circular boundary.
But the boundary ribbon does not remain circular throughout deformation; hence, \eq{f_bend} is only accurate for small $\delta_{x}$ values. 
As the displacement $\delta_{x}$ grows the boundary ellipse becomes flat, increasing its aspect ratio $a/b$ and decreasing the radius of curvature at the ends of its major axis. 
As such, the actual bending force required to deform the boundary is higher than the force estimated by \eq{f_bend} (see our simulations in Appendix~\ref{sec:bend}).

In order to account for the inaccuracy of \eq{f_bend} at large displacement values, we next introduce the stretching force $F_{stretch}$.
Consider the ellipse when fully extended. At this extreme the semi-minor axis converges to zero (i.e., $b\rightarrow 0$), and the boundary ribbon becomes two straight, parallel ribbons with length $\pi r$.
Increasing $\delta_x$ any further must cause the ribbons to stretch --- they are already fully extended.
Accordingly, when displacement $\delta_{x} > r(\pi - 2)$, we set the bending force $F_{bend} = 0$ and compute the stretching force for the ribbons using Hooke's law:
\begin{equation}
    F_{stretch} = \frac{2EA \delta_{l}}{\pi r} \label{eq:f_stretch}
\end{equation}
Here $A$ is the area of cross-section and the change in length $\delta_{l} = \delta_{x} - r(\pi - 2)$.

Our final step for the mechanics of the boundary ribbon is to combine the bending and stretching forces.
In practice, the boundary is pulled using a rigid attachment of width $2b_{min}$; hence, once the semi-major axis $b$ is equal to $b_{min}$, the boundary ribbon can no longer bend.
We leverage this critical point to determine whether to apply \eq{f_bend} or \eq{f_stretch}.
Overall, we model the total force required to deform the boundary ribbon as:
\begin{equation}
F_{boundary} = 
\begin{cases}
    F_{bend} & b \leq b_{min}\\
    F_{stretch} & b > b_{min}
\end{cases}
\label{eq:f_boundary}
\end{equation}
We note that \eq{f_boundary} is necessarily a \textit{lower bound} on the actual force.
In practice, the boundary may start stretching even before it flattens to $b_{min}$, and the effective radius of the boundary ribbon is not consistently $r$.

\subsection{Discrete Ribbons Bending}\label{sec:discrete}

The boundary ribbon forms the edge of the spoon, and is critical for pinching behaviors.
But the base of the spoon is composed of multiple discrete ribbons --- and these ribbons have an essential role in the shape of Kiri-Spoon.
Interestingly, the discrete ribbons oppose the deformation of the boundary ribbon.
Referring to \fig{design1} and \fig{design2}, we note that the discrete ribbons are parallel to the minor axis of the boundary layer.
When the kirigami sheet is actuated, these discrete ribbons must \textit{bend} in order for the boundary ribbon to elongate.
Consider \fig{dynamics}(a) and \fig{dynamics}(b): increasing displacement $\delta_x$ decreases the semi-minor axis $b$.
This in turn compresses the discrete ribbons axially, causing each ribbon to bend into an arch.
Below we derive the shape of these arches, as well as the forces the boundary ribbon applies to the discrete ribbons.

\p{Geometry} Building on our early work \citep{keely2024kiri}, we model the arch formed by a discrete ribbon as a catenary:
\begin{equation}
    l_{ribbon} = 2\alpha\sinh(d_{y} / 2\alpha) \label{eq:catenary_length}
\end{equation}
\begin{equation}
    d_{z}^{2} + 2\alpha d_{z} - (l_{ribbon}/2)^{2} = 0 \label{eq:catenary_depth}
\end{equation}
Here $l_{ribbon}$ is the undeformed length of the discrete ribbon, $d_{y}$ is the distance between its endpoints, $d_{z}$ is the maximum depth at its center, and $\alpha$ is a parameter that defines its shape.
Because each discrete ribbon has a different length, these values will vary along the kirigami structure.

\p{Force} Equipped with this geometry, we next seek to determine how much tensile force is required to bend the discrete ribbons (i.e., $F_{discrete}$). 
Modeling $F_{discrete}$ is necessary to determine the overall force required to actuate Kiri-Spoon: in order to elongate the kirigami sheet, we must overcome the resistance of the discrete ribbons.
For the $i$-th discrete ribbon, let $P_i$ be the axial force exerted by the boundary ribbon.
Similarly, let $\mathcal{P}_i = -P_i$ be the equal and opposite force that $i$-th discrete ribbon applies back to the boundary.
As shown in \fig{dynamics}(b), these forces lie in the same plane as the boundary layer. 

We can compute the resistance force $\mathcal{P}_i$ as a function of the discrete ribbon's geometry and material properties.
When the kirigami structure is actuated, the boundary ribbon applies a force at both ends of the $i$-th discrete ribbon.
We estimate this force by treating each half of the discrete ribbon as a cantilever beam \citep{gere2009mechanics}:
\begin{equation}
    P_i \sin \phi = \frac{3EId_{z}}{(l_{ribbon}/2)^{3}} \label{eq:f_discrete1}
\end{equation}
Within this equation the maximum deflection of the cantilever beam is set equal to the depth of the catenary.
Here $\phi$ is the approximate angle between the applied force $P$ and the discrete ribbon:
\begin{equation} \label{eq:f_discrete2}
    \phi = \tan^{-1}\left(\frac{d_{z}}{d_{y}/2}\right)
\end{equation}
Combining \eq{f_discrete1} and \eq{f_discrete2} enables us to solve for $P_i$, the compressive force applied by the boundary ribbon.
The resistance force for the $i$-th discrete ribbon is simply the equal and opposite force: i.e., $\mathcal{P}_i = -P_i$.

We now have the resistance force from one discrete ribbon; our final step is to estimate the cumulative force $F_{discrete}$ needed to bend all the discrete ribbons.
Let $n_{d}$ be the number of ribbons.
We assume that these ribbons are spaced uniformly along the major axis of the kirigami sheet.
As shown in \fig{dynamics}(b), the contribution of each discrete ribbon to $F_{discrete}$ depends on the length of the ribbon (as described in \eq{f_discrete1}), the distance between the discrete ribbon and the point of tensile load (i.e., the moment arm), and the angle $\theta$ between the tensile force direction and the boundary. 
As such, the force is directly proportional to the moment arm and inversely proportional to the ribbon length.
For example, the discrete ribbon at the center (i.e., $i=\lceil n_{d}/2 \rceil$) has the greatest length, resulting in the smallest opposing force $\mathcal{P}_{i}$ but also the largest moment arm. 

We provide detailed calculations for these effects in Appendix~\ref{sec:discrete_component}.
In conclusion, we find that the total tensile force due to resistance from all discrete ribbons is given by:
\begin{equation}
    F_{discrete} = 2\sum_{i=1}^{ \lceil n_{d}/2 \rceil} \frac{P_{i}}{\tan \theta} \left(\frac{i}{\lfloor n_{d}/2 \rfloor + 1}\right) \label{eq:f_discrete}
\end{equation}
Intuitively, in order to bend the discrete ribbons the actuator must apply a force greater than $F_{discrete}$.

\begin{figure}[t]
    \begin{center}
        \includegraphics[width=1\linewidth]{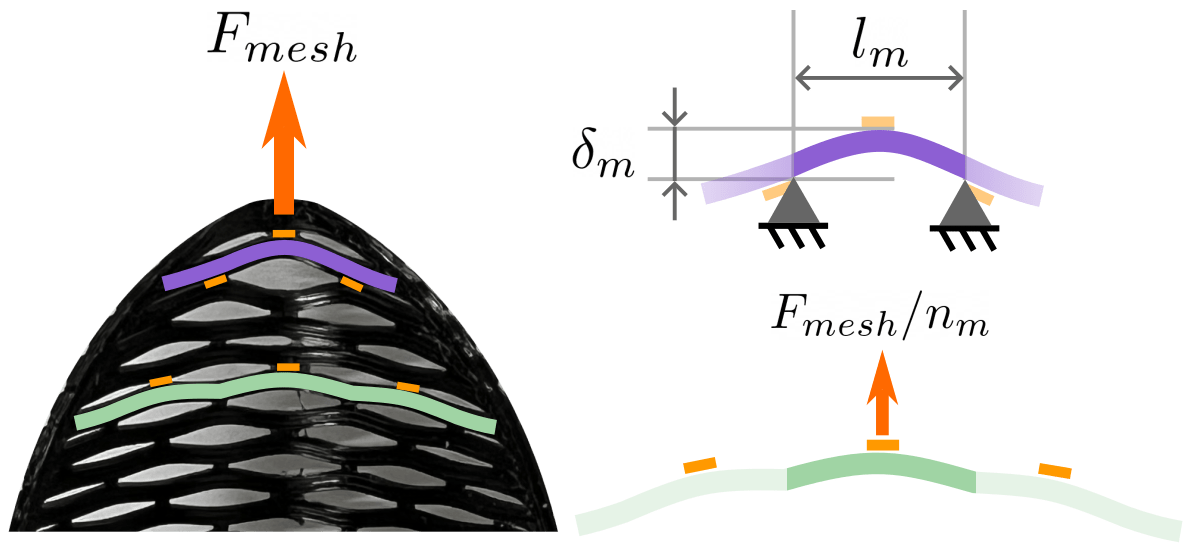}
        \vspace{-3 ex}
        \caption{Dynamics of the mesh and discrete ribbons under tensile load. $F_{mesh}$ is the additional tensile force component needed to deform the kirigami sheet due to the mesh ribbons. The tensile force is equally divided into the mesh ribbons. Each mesh ribbon bends a section of the discrete ribbon. For example, the green ribbon is connected to three mesh ribbons. Therefore, the load on its central section is $F_{mesh}/3$ and the corresponding deflection is $\delta_{m, 1}/3$. The total deflections along the central ribbon must equal the total displacement $\delta_{x}$.}
        \label{fig:mesh}
        \vspace{-2 ex}
    \end{center}
\end{figure}

\begin{figure*}[t]
    \begin{center}
        \includegraphics[width=1\textwidth]{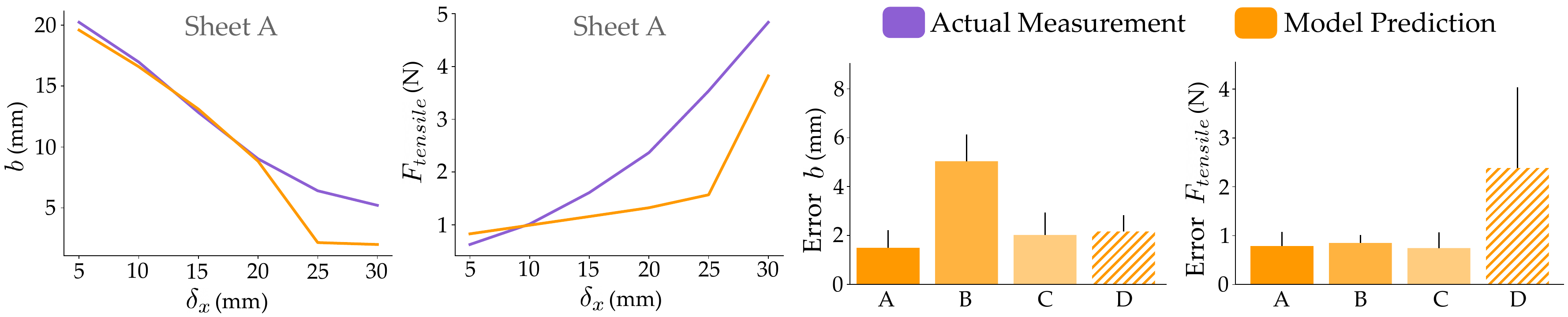}
        \vspace{-3 ex}
        \caption{Results of validation experiments in Section~\ref{sec:validation}. (Left) The half-width $b$ (semi-minor axis) and total tensile forces $F_{tensile}$ predicted by our model for sheet A. The predicted and measured widths closely align up to a displacement of $\delta_{x} = 20$, while the predicted forces underestimate the actual tensile force required to deform the kirigami sheet. Our predictions deviate from the actual measurements because we do not account for the torsion or stretching of the boundary ribbon before reaching the minimum width of $b_{min}$. (Right) For all sheets except B, the mean absolute error in the predicted half-width is approximately $2$ millimeters (mm). Moreover, the mean absolute error in the predicted forces is less than $1$ Newton (N) for all sheets except D. Note that sheet D has a significantly higher Young's modulus, leading to larger deformation forces. We use sheet A in our robot experiments.}
        \label{fig:model_results}
        \vspace{-2 ex}
    \end{center}
\end{figure*}

\subsection{Mesh Ribbons Resistance} \label{sec:M3}

So far we have derived the mechanics of the boundary ribbon and discrete ribbons. 
The only remaining element for our kirigami sheet is the mesh (see \fig{design1} and \fig{design2}).
Here we will only focus on the discrete mesh ribbons, since they are sufficient for most foods and can significantly impact kirigami mechanics.
Within our design the mesh ribbons are used to interconnect the discrete ribbons.
At alternating points a short mesh ribbon is placed between neighboring discrete ribbons, ultimately forming a grid-like pattern.
In practice, the mesh ribbons help the Kiri-Spoon retain its shape while preventing small foods from slipping through the gaps.
However, these mesh ribbons also increase the amount of force required to actuate our kirigami sheet.

\p{Force} We treat each of the mesh ribbons as an inextensible beam.
Hence, we do not need to consider the geometry of these ribbons --- just the effect that they have on actuator force.
As shown in \fig{mesh}, when the kirigami sheet is extended the mesh ribbons cause the attached discrete ribbons to bend.
Specifically, each mesh ribbon bends a section of the discrete ribbon between prior and subsequent meshes.
We consider these sections to be a simply supported beam with a center load of $Q$ \citep{gere2009mechanics}:
\begin{equation}
    Q = \frac{48EI\delta_{m}}{l_{m}^{3}} \label{eq:f_cantilever}
\end{equation}
Here $l_{m}$ is the length of the section and $\delta_{m}$ is the maximum deflection of that section.
The length $l_{m}$ is determined by the sheet's geometry (and therefore known).
On the other hand, the deflection $\delta_{m}$  is unknown, and must be calculated as described below.
Intuitively, computing $Q$ is important because it captures how tensile force is propagated along the mesh.
Suppose that the first discrete ribbon ($i=1$) is connected to the boundary layer via one central mesh (as shown in \fig{mesh}). 
Then the overall resistance force caused by the mesh ribbons is equal to $Q_{i=1}$, i.e., \eq{f_cantilever} evaluated for the first discrete ribbon.
We therefore set $F_{mesh} = Q_1$, where $F_{mesh}$ is the total tensile force caused by the mesh ribbons.

Unfortunately, in order to apply \eq{f_cantilever} and find $Q_1$ we must identify the deflection $\delta_{m,1}$ of the first section.
Because this first section is connected by mesh ribbons to a second section, and so on, finding $\delta_{m,1}$ becomes a recursive process.
Consider the connection between the second and third discrete ribbons.
To compute the force exerted by the central ribbon on the second discrete ribbon, i.e., $Q_{2}$, we assume that $F_{mesh}$ is equally distributed into the mesh ribbons.
More formally, if there are $n_{m, i=2}$ mesh ribbons connecting the second and third discrete ribbons, the tensile load on the central section is $Q_{i=2} = F_{mesh} / n_{m, i=2}$ and the deflection in the second discrete ribbon is $\delta_{m, i=2} = \delta_{m, i=1} / n_{m, i=2}$. More generally, the deflection in the $i$-th discrete ribbon becomes:
\begin{equation}
    \delta_{m, i} = \frac{\delta_{m, 1}}{n_{m, i}}
\end{equation}
We also recognize that the total deflection along the center for all discrete ribbons must equal the total deformation of the kirigami sheet in the direction of the tensile force:
\begin{equation}
    \delta_{x} = \sum_{i}^{n_{d}} \delta_{m, i} = \delta_{m, 1} \sum_{i}^{n_{d}} \frac{1}{n_{m, i}} \label{eq:mesh_deform}
\end{equation}
By leveraging \eq{mesh_deform} we can complete our recursive reasoning and calculate $\delta_{m,1}$.
For any $\delta_{x}$, we first obtain $\delta_{m, 1}$ using \eq{mesh_deform}, and then apply $\delta_{m, 1}$ to calculate $Q_1$ using \eq{f_cantilever}.
To overcome the resistance of the mesh ribbons and extend the kirigami sheet, the actuator must apply a force greater than $F_{mesh} = Q_1$.

\subsection{Summary}

We conclude our mechanics analysis by combining each of the effects covered in Sections~\ref{sec:boundary}, \ref{sec:discrete}, \ref{sec:M3}.
By adding the forces applied by the boundary ribbon, discrete ribbons, and mesh ribbons, we reach a lower bound on the total tensile force $F_{tensile}$ needed to actuate Kiri-Spoon.
\begin{equation}
    F_{tensile} = F_{boundary} + F_{discrete} + F_{mesh} \label{eq:f_total}
\end{equation}
The force $F_{tensile}$ is a lower bound because of the approximations necessary to capture interconnected ribbon mechanics.
As we will show in the following validation tests, however, this is a \textit{tight} lower bound with errors less than $1$~N across different kirigami materials and geometries. 
Typical micro linear actuators that apply forces up to $50$ N can easily compensate for this error.
% which is less than $20\%$ of the tensile force required to fully actuate kirigami sheets of different dimensions.

\subsection{Validation Experiments} \label{sec:validation}

Our theoretical model simplifies the calculation of the force required to actuate the kirigami sheet and the dimensions of the resulting kirigami structure. 
We now test this model to validate whether our approximations hold in practice, and to see if other designers can apply our model to develop their own Kiri-Spoons.
For validation experiments we created four kirigami sheets with varying geometries and material properties (see Table~\ref{tab:sheet}), and compared our model predictions to the actual forces and sheet dimensions during testing.

\begin{table}[t]
\caption{Kirigami sheet properties}
\label{tab:sheet}
\resizebox{\columnwidth}{!}{%
\begin{tabular}{c|c|c|c|c}
\hline
Sheet & Material & Radius (mm) & Thickness (mm) & Ribbon width (mm) \\ \hline
\textbf{A} & TPU & 22.24 & 1 & 1 \\
B & TPU & 22.24 & 1.5 & 1 \\
C & TPU & 16.68 & 1 & 0.75 \\
D & PET & 22.14 & 0.25 & 0.8 \\ \hline
\end{tabular}
}
\vspace{-2 ex}
\end{table}

We designed sheet A based on stakeholder feedback, using a soft TPU material and a radius tuned for user comfort while eating. 
To investigate how the thickness of the kirigami sheet affects the actuation force, we increased its thickness to create sheet B. 
Next, we varied the radius of sheet A while maintaining the same sheet thickness and the number of discrete ribbons to produce sheet C. 
Finally, we fabricated sheet D using the PET material which has been used in previous kirigami-based grippers~\citep{yang2021grasping}.
This PET material was described by our stakeholders as ``too stiff'' to be comfortable for eating.

We tested each of these sheets by attaching one end to a calibrated force sensor, and the opposite end to a linear screw.
The screw was initially set to the undeformed length of the sheet and then actuated in increments of $\delta_{x} = 5$ millimeters (mm).
Moving the screw displaced the boundary ribbon perpendicular to the discrete ribbons, causing the kirigami structure to deform into a $3$D bowl. 
For each increment of displacement, we recorded the force measured by the sensor and the width of the deformed boundary ribbon.

\fig{model_results} summarizes our results. 
On the left, we illustrate our model predictions for sheet A. 
Using \eq{perimeter}, we can accurately estimate the half-width (i.e., the semi-minor axis $b$) of the boundary up to $\delta_{x} = 20$ mm. 
For larger displacements, the boundary starts to stretch, increasing its length and causing our model --- which assumes a constant boundary perimeter --- to underestimate the width. 
Our model also closely approximates the tensile force needed to deform the kirigami sheet. 
The predicted force within our results is lower than the measured force primarily because we do not account for stretching until the boundary reaches its minimum width, as defined in \eq{f_boundary}.

On the right of \fig{model_results} we plot the mean absolute error in the model predictions for each different kirigami sheet. 
The average error in the predicted half-widths is only $2$ mm for all sheets except B. 
Our rational here is that --- as the thickness of the sheet increases --- the discrete ribbons apply a greater opposing force on the boundary, increasing its width. 
However, our model does not account for this force when estimating the boundary dimensions. We also find that the average error in the predicted forces is less than $1$ Newton (N) for all sheets except D.
This is because sheet D is made of PET, which has a Young's modulus ($E$) of $3.57$ GPa --- approximately $200$ times that of 3-D printed TPU --- which has $E = 14.77$ MPa. 
As a result, the forces required to deform sheet D are significantly higher than the TPU sheets, leading to higher absolute prediction errors.
Yet, across all sheets, the average error in our predictions is less than 25\% of the maximum force required to actuate the sheets fully.

\p{Summary and Personalization} Our validation tests suggest that the proposed mechanics model is an accurate lower bound (i.e., $<1$N error) for the tensile force required to actuate kirigami sheets of varying thicknesses, sizes, and materials.
In the subsequent experiments --- including Sections~\ref{sec:auto_experiment}, \ref{sec:userstudy}, and \ref{sec:finalstudy} --- we will apply a Kiri-Spoon with sheet A.
This particular sheet was designed based on the subjective feedback of our stakeholders, and we applied the theoretical forces calculated for this sheet to choose a suitable lead screw and motor for actuating our Kiri-Spoon.

However, we recognize that each individual user may prefer kirigami sheets of different size, thickness, or material, and this preference may vary based on the types of foods that user consumes.
To account for this personalization, designers can manufacture a customized sheet by starting with our CAD models and then modifying the dimensions according to their end-user's needs. 
For example, a wider kirigami sheet may be more suitable for stakeholders who prefer to take bigger bites of food --- but this change would also require higher actuation forces. Designers can leverage our mechanics model to compute the maximum force required to actuate the kirigami sheet fully, and then select a linear actuator that reliably supplies that force.
\begin{figure*}[t]
    \begin{center}
        \includegraphics[width=1\textwidth]{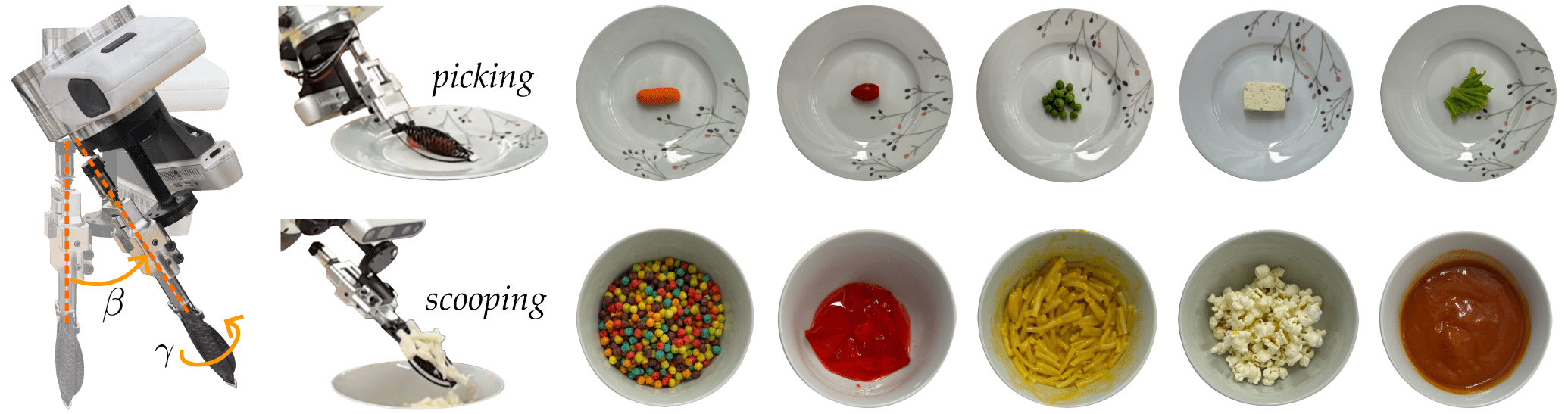}
        \vspace{-3 ex}
        \caption{Experimental setup for the autonomous tests in Section~\ref{sec:auto_experiment}. (Left) Position and orientation of Kiri-Spoon during autonomous acquisition. When picking foods from a plate, the flexible hoop and kirigami sheet bend to align with the orientation of that plate. Upon reaching the target position, we rapidly increase the curvature of the kirigami sheet to firmly grasp the desired food. In contrast, Kiri-Spoon maintains a spoon-like curvature when scooping food from a bowl. (Right) Foods used in our acquisition experiments. For picking, we include round foods of different sizes, i.e., carrots, cherry tomatoes, and peas. We also test with soft and slippery foods like silken tofu and flat foods like lettuce. For scooping, we include dry foods like cereal and popcorn, sticky foods like macaroni and cheese, slippery foods like jello, and liquid foods like tomato soup.}
        \label{fig:auto_acquire}
        \vspace{-2 ex}
    \end{center}
\end{figure*}

\begin{figure*}[t]
    \begin{center}
        \includegraphics[width=0.95\textwidth]{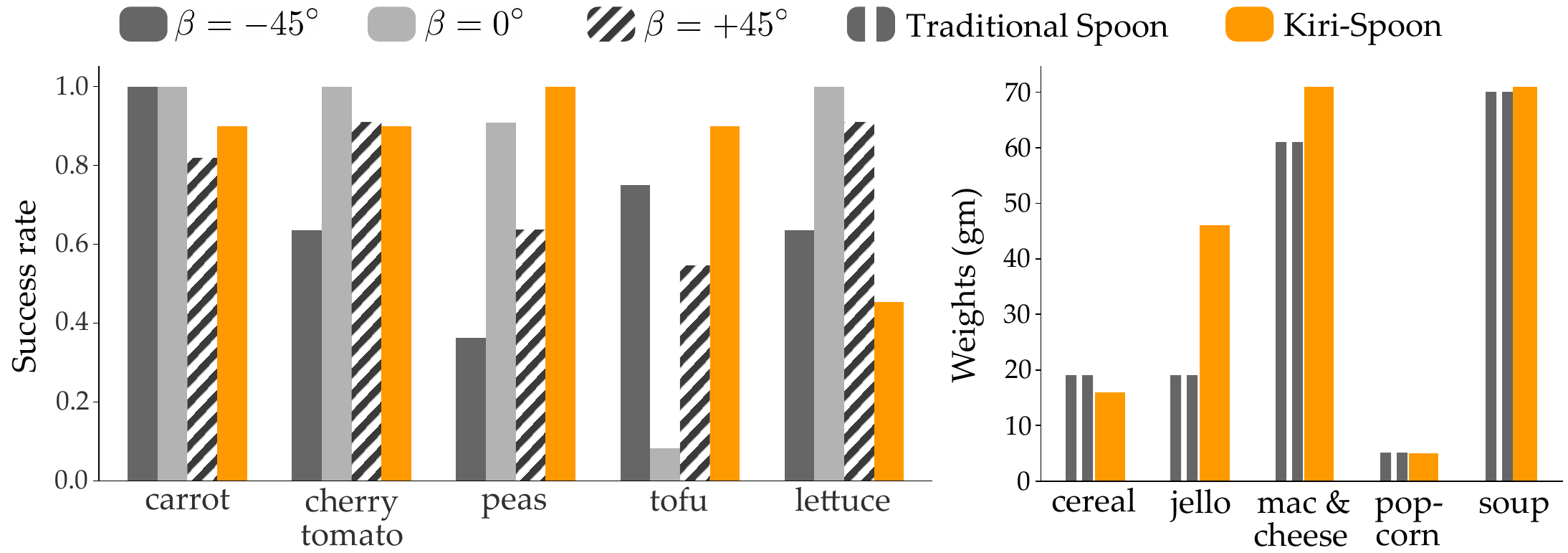}
        \vspace{-1 ex}
        \caption{Results for autonomous acquisition tests in Section~\ref{sec:auto_experiment}. (Left) Kiri-Spoon successfully picks round foods such as carrots, tomatoes, peas, and tofu, but struggles to pick flat foods like lettuce as compared to a traditional fork. While the pitch of the fork needs to be changed according to the shape and hardness of the food, Kiri-Spoon is easier to control because it does not require any pitch adjustment. (Right) Kiri-Spoon scoops the same amount of dry and sticky foods as a traditional spoon. Across both tasks, Kiri-Spoon outperforms the traditional utensils in acquiring slippery foods like jello and tofu.}
        \label{fig:auto_results}
        \vspace{-3 ex}
    \end{center}
\end{figure*}

\section{Autonomous Acquisition} \label{sec:auto_experiment}

We have presented our design for Kiri-Spoon, and modeled the mechanics of this kirigami utensil.
Next, we begin to test Kiri-Spoon's capabilities: can our utensil created specifically for robot arms effectively and easily acquire a wide range of bite-sized foods?
In this section we compare a robot arm equipped with traditional forks and spoons to the same robot using a Kiri-Spoon.
To measure the best-case performance of each system, we autonomously control the robot arm using a state-of-the-art algorithm for robot-assisted feeding.
Our objective is to evaluate the mechanical advantage offered by Kiri-Spoon; i.e., whether adding Kiri-Spoon will make the overall system more effective at picking up diverse foods such as carrots, peas, tofu, cereal, jello, and soup.
We hypothesize that Kiri-Spoon's ability to encapsulate the bites will enable it to more successfully pick soft and slippery foods, which often slide off spoons and forks. 
Moreover, while regular forks need to be precisely oriented for successful acquisition, we expect the Kiri-Spoon to be equally effective while maintaining a constant orientation because of its compliant structure.

\subsection{Task and Experimental Setup}
Images of the attached Kiri-Spoon and target foods are shown in \fig{auto_acquire}.
We mounted each feeding utensil in place of the end-effector for a 7-DoF Franka Emika robot arm \citep{frankaemika},
To detect foods, we also mounted an Intel RealSense D435 camera on the robot's wrist.
The robot used its utensil and camera to autonomously perform two types of acquisition tasks: (i) \textit{picking} food off a plate and (ii) \textit{scooping} food from a bowl. 

For \textit{picking}, we autonomously controlled the acquisition motion using a pre-trained \textit{Skewering-Position-Action Network (SPANet)} \citep{feng2019robot, gordon2023towards}.
This network outputs the target position $(x, y, z)$, roll $\gamma$, and pitch $\beta$ of robot's fork based on the location and orientation of the detected food. 
Similar to prior work, we discretized the pitch into three angles: $-45^{\circ}$ (\textit{tilted backward}), $0^{\circ}$ (\textit{vertical}), and $+45^{\circ}$ (\textit{tilted forward}). 
While we adjusted the pitch of the traditional fork based on the SPANet output, we maintained a constant pitch of $\beta = 45^{\circ}$ for the Kiri-Spoon (see \fig{auto_acquire}, Left).
For \textit{scooping}, we pre-programmed a fixed motion pattern for both the traditional spoon and the Kiri-Spoon. 
This motion pattern was tuned in offline experiments to maximize the success rate of the traditional spoon. 
In addition to controlling the robot's motion, we used an Arduino Uno board to autonomously actuate Kiri-Spoon to a pre-defined spoon-like shape in the scooping task and to a high curvature state in the picking task after reaching the food.

\subsection{Independent Variables}
We compared the acquisition performance of the Kiri-Spoon to a traditional fork in the picking task and to a traditional spoon in the scooping task. 
Inspired by related works \citep{feng2019robot, sundaresan2023learning2, liu2024imrl}, we selected five foods of varying size, shape, hardness, and consistency for each task to test the Kiri-Spoon in acquiring diverse foods.
The robot manipulated its traditional fork to try to pick up carrots, cherry tomatoes, peas, silken tofu, and lettuce bites (\fig{auto_acquire}, Top).
Similarly, the robot leveraged its traditional spoon to try to scoop cereal, macaroni and cheese, jello, popcorn, and tomato soup (\fig{auto_acquire}, Bottom).
After completing trials with the traditional utensils, the robot then attempted to iteratively acquire, carry, and release all of these foods using a single Kiri-Spoon and its food-safe kirigami sheet.
The robot's control algorithm remained constant across the experiment.

\subsection{Dependent Variables}
The robot arm attempted to pick or scoop each food $10$ times using both the traditional utensils and the Kiri-Spoon. 
In the picking task we recorded the percentage of \textbf{successful attempts}, while in the scooping task we measured the total \textbf{weight of food} collected over $10$ attempts.
We considered an individual acquisition attempt to be successful if the food stayed on the utensil for $5$ seconds after it was picked. For a fair comparison of the weight of food acquired, we ensured that the spoon and Kiri-Spoon had similar volumes.

\subsection{Results}
Our experimental results are summarized in \fig{auto_results}.
These results should be viewed as the current \textit{best case} performance for robot arms using forks and spoons; we applied assistive eating algorithms specifically designed for these utensils, and tuned the experimental setup to maximize their performance.
But even in this best-case setting, we found that Kiri-Spoon matched or outperformed the traditional utensils across most foods.
Kiri-Spoon picked up all types of foods besides lettuce with a success rate of more than $80\%$.
Lettuce was a failure case for Kiri-Spoon: its flat, thin surface stuck to the plate, and Kiri-Spoon was unable to get purchase to pinch the morsel.
On the other hand, the robot was usually able to pick up lettuce with a traditional fork.
Fundamentally, Kiri-Spoon employs a different grasping mechanism than standard forks --- while the fork skewers the food to hold it using friction, Kiri-Spoon encapsulates the food within its kirigami bowl. 
It is therefore challenging for Kiri-Spoon to pick up items that lay flat on the plate, as well as foods that are larger than the radius of its kirigami sheet. 

By contrast, Kiri-Spoon's mechanical intelligence enables the robot to successfully grasp a diverse set of foods without ever changing end-effector orientation.
When robots use standard forks, prior works and our results demonstrate that the fork's orientation and manipulation is crucial for acquisition \citep{feng2019robot, gordon2023towards, liu2024imrl, tai2023scone}.
For example, to pick up round and cylindrical foods like carrots, tomatoes, and peas, the traditional fork needs to have a pitch $\beta = 0^{\circ}$ so that its tines enter the food vertically and the item does not roll away.
But when the robot is using a fork to pick up soft and slippery foods like silken tofu, the tines need to be angled to ensure that the food stays on the fork. 
In contrast to the fork --- which the robot arm needs to carefully manipulate --- the robot can hold Kiri-Spoon at a constant orientation.
In \fig{auto_results} Kiri-Spoon successfully encapsulates and carries both food types while keeping a constant pitch of $\beta=45^{\circ}$.
This capability can especially be useful in constrained environments where reaching the desired angle for a fork can cause collisions with objects on the dining table. In such scenarios, a robot leveraging Kiri-Spoon can move to an orientation that avoids collisions and use its flexible hoop to reach and successfully acquire the food. 

Now focusing on the scooping task, we find that Kiri-Spoon functions similar to traditional spoons for both dry foods, such as cereal and popcorn, and sticky foods, like macaroni and cheese. 
However, the two utensils perform differently for slippery foods (e.g, jello).
When handling slippery foods with traditional spoons the jello can easily slide off the utensil, particularly if the robot moves quickly or with the wrong angle.
Kiri-Spoon is more effective here because it encapsulates the slippery jello pieces within its kirigami structure, preventing them from accidentally falling.
We note that for all the foods except soup we leveraged a kirigami sheet with a discrete mesh (see \fig{design2}, Top).
This was sufficient to acquire and carry foods with solid or viscous components.
But for the soup --- which was a liquid --- we applied a kirigami sheet with a continuous mesh (see \fig{design2}, Bottom).
This continuous mesh was necessary to keep the soup from flowing out of the bottom of Kiri-Spoon.
Across autonomous testing we did not find that the type of mesh had a noticeable affect on our Kiri-Spoon performance; both mesh types can be used interchangeably.
We typically prefer the discrete mesh because it is easier to manufacture and actuate.
While kirigami sheets with a discrete mesh can be directly 3D printed, manufacturing the continuous mesh requires using an ellipsoidal mold to shape the membrane before the sheet can be printed onto it.

\p{Summary} Our autonomous tests highlight three mechanical advantages of Kiri-Spoon.
First, Kiri-Spoon is a single feeding utensil that can function both as a fork (picking carrots from a plate) and as a spoon (scooping soup from a bowl).
This utility is practically important for assistive feeding scenarios since changing utensils means that the robot arm must switch its end-effector.
Instead of detaching a fork to mount a spoon --- or \textit{vice versa} --- here the robot arm can stick with just a single utensil.
Second, we find that the robot arm can effectively leverage Kiri-Spoon like a fork without needing to change its orientation.
This makes the arm's manipulation task easier: the system does not need to tune the end-effector angle to pick up different types of foods.
Finally, while the Kiri-Spoon struggles to consistently pick large and flat foods like lettuce, it outperforms traditional utensils in handling slippery foods such as jello and tofu.
This matches our expectations --- Kiri-Spoon can enclose foods within its kirigami structure, thereby preventing morsels from unintentionally falling during interaction.

\begin{figure*}[t]
    \begin{center}
        \includegraphics[width=1\textwidth]{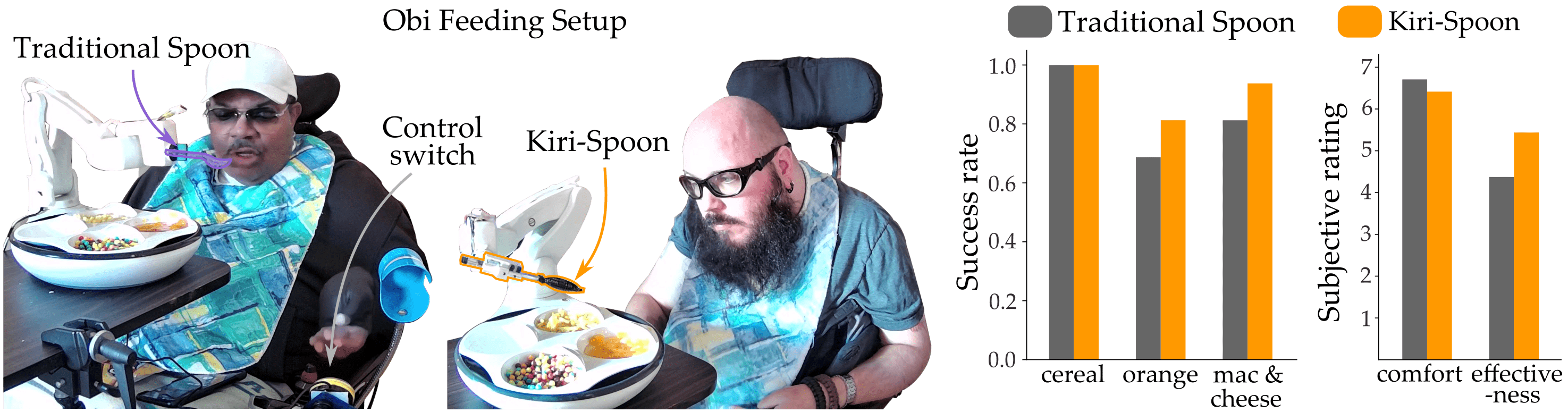}
        \vspace{-3 ex}
        \caption{Experimental setup and results from our second round of stakeholder tests in Section~\ref{sec:userstudy}. (Left) Residents of The Virginia Home interacting with the Obi feeding device and scooping food using a traditional spoon and Kiri-Spoon. (Right) Objective and subjective results across $N=4$ adults who require assistance when eating. Kiri-Spoon had a slightly higher success rate than the traditional spoon when picking canned oranges (i.e., a slippery food) and macaroni and cheese (i.e., a sticky food). Both utensils were equally effective at picking cereal (i.e., a dry food). Overall, users perceived the Kiri-Spoon to be almost as comfortable as the traditional spoon while being more effective in picking and securely carrying the food to their mouth.}
        \label{fig:virginia_results}
        \vspace{-3 ex}
    \end{center}
\end{figure*}

\section{User Studies with Participants with Disabilities} \label{sec:userstudy}

Our autonomous study in Section~\ref{sec:auto_experiment} demonstrated that Kiri-Spoon can effectively acquire, carry, and release a diverse set of bite-sized foods.
This functionality suggests that --- from the robot's perspective --- Kiri-Spoon is an advantageous utensil.
But what about the human's perspective?
In this section we interact with stakeholders who require assistance during eating to assess the comfort and performance of Kiri-Spoon.
The purpose of these studies is to determine whether Kiri-Spoon addresses the needs of its intended users.

We performed this experiment across two sessions at The Virginia Home \citep{virginiahome}.
During both sessions the Kiri-Spoon was attached to an Obi \citep{obi}, a table-top robot arm and bowl system that is commercially available for assistive feeding (see \fig{virginia_results}, Left).
By default, the Obi is equipped with a traditional spoon that it employs to scoop foods from the bowl and carry them to the user's mouth.
We compared this default setup to an Obi with our Kiri-Spoon mounted in place of the standard spoon.
Our first experimental session focused on identifying the design characteristics necessary for a comfortable Kiri-Spoon (also see Section~\ref{sec:design}).
Given this guidance from the stakeholders, in our second experimental session we returned with a finalized Kiri-Spoon.
Here occupational therapists and $N=4$ residents with mobility limitations compared our finalized Kiri-Spoon to the traditional spoon, and we collected objective and subjective results from their interactions.
Both of these user studies followed the same general procedure.
We have already discussed the design considerations (i.e., the outcomes of the first session) in Section~\ref{sec:design}.
Accordingly, here we will focus on our experimental procedure and results from the second visit with stakeholders\footnote{For videos of this study and our other experiments, please see \url{https://youtu.be/ZJpQREdTz80}}.

\subsection{Experimental Setup}
In both visits we tested the Kiri-Spoon attached to an \cite{obi}, a medical device specifically designed for assistive feeding. 
The Obi consists of a $6$-DoF robot arm and utensil with four bowls for storing food (see \fig{virginia_results}, Left). 
Participants were able to control Obi using two switches: one to change which bowl the robot will scoop from, and the other to scoop food from that bowl and bring it to the human's mouth. 
The simplicity of the control interface enables users with mobility impairments to feed themselves independently.
We positioned the control switches according to the mobility range of each participant.
In addition, we integrated Kiri-Spoon within the robot's control circuit so that it automatically closed its kirigami sheet when scooping the food, and then opened that kirigami sheet after the morsel was brought to the human's mouth.
The additional weight of the Kiri-Spoon had no discernable impact on Obi's ability to scoop and lift the food to the user's mouth.

\subsection{Participants}
During both visits to \cite{virginiahome}, we interacted with $N = 4$ residents living with upper-limb mobility impairments. 
Three of these residents participated in both sessions but the fourth participant was different in each session.
All participants were adult men aged $40 \pm 10.5$ years, and they each provided informed consent in accordance with Virginia Tech University guidelines (IRB $\# 22$-$308$).
Three of the participants identified themselves as quadriplegic, and the two other participants self-identified as having limited arm mobility or arm spasms.
All of these users reported that they depend upon a caregiver everyday in order to eat their meals.
The leading occupational therapist from The Virginia Home supervised the sessions to ensure the participant's safety.

\subsection{Independent Variables}
The Obi is pre-programmed to execute scooping motions using a fixed trajectory, and this system cannot perform picking tasks. 
Therefore, in this study we only compared \textbf{Kiri-Spoon} to a \textbf{traditional spoon}.
Participants used the Obi equipped with both utensils to scoop and eat three different types of foods.
In accordance with the participant's preferences, we tested a dry food (cereal), a sticky food (macaroni and cheese), and a slippery food (canned oranges).
The experiments followed a within-subjects design: all participants attempted to scoop and eat each type of food four times with the traditional spoon, and four times with Kiri-Spoon.
The order of presentation was balanced so that half of the participants started with Kiri-Spoon.

\subsection{Dependent Variables}
We collected objective and subjective measures for each utensil.
To assess objective performance, we recorded the number of scooping attempts where the Obi successfully acquired food.
We report this metric as \textit{Success Rate}, defined as the number of successful acquisitions divided by the number of attempted acquisitions.
To better assess the stakeholder's subjective perception of the system, we asked each participant to answer a survey after using both utensils. 
The survey items are listed in Table~\ref{tab:questions1}.
We grouped these items into two scales: \textbf{Comfort} and \textbf{Effectiveness}. Items in the comfort scale assessed the perceived comfort, intuitiveness, and safety of each feeding utensil.
Items in the effectiveness scale measured the perceived ability of the utensils to successfully acquire, carry, and transfer morsels without spilling. 
Finally, at the end of each session we asked participants to indicate their preferred feeding utensil (forced-choice question), and provide open-ended responses about the advantages and limitations of Kiri-Spoon.

\begin{table}[t!]
\centering
\caption{Survey for participants living with physical disabilities (Likert scales with $7$-options)}\label{tab:questions1}
\resizebox{\linewidth}{!}{%
\begin{tabular}{l}
\hline\\[-1.5 ex]

\textbf{Comfort:}\\
Q1. This utensil was comfortable to use.\\
Q2. This utensil was jarring to use and not comfortable.\\
% \vspace{1 ex}\\

% \textbf{Intuitive:}\\
Q3. This utensil was simple to figure out how to eat off of.\\ 
Q4. It was difficult to understand how to eat out of this utensil.\\    
% \vspace{1 ex}\\ 

% \textbf{Safe:}\\
Q5. I felt safe and comfortable being fed using this utensil.\\ 
Q6. I felt apprehensive being fed by this utensil.
\vspace{1 ex}\\ 

\textbf{Effectiveness:}\\
Q7. The utensil picked up the desired food consistently. \\
Q8. This utensil was unable to get a lot of food.\\
% \vspace{1 ex}\\ 

% \textbf{Secure:}\\
Q9. I felt like this utensil kept the food secure until it reached me.\\ 
Q10. I was often worried the food would fall off the utensil.
\vspace{0.75 ex}\\ 
\hline
\end{tabular}%
}
\vspace{-1 ex}
\end{table}

\subsection{Results}
Our results from experiments with the finalized Kiri-Spoon are summarized in Figure~\ref{fig:virginia_results} (Right).
For these results, we note that the sample size ($N=4$) was not sufficient to reliably perform statistical tests.

Objectively, both utensils had similar success rates.
The spoon and Kiri-Spoon acquired cereal in all trials, and the Kiri-Spoon was marginally better at scooping slippery oranges as well as sticky macaroni and cheese.
Subjectively, users perceived the regular spoon to be comfortable to use and eat from. 
This outcome was expected given people's familiarity with traditional spoons. 
Interestingly, participants rated the Kiri-Spoon to be almost as comfortable as the regular spoon despite their lack of familiarity with this device.
We also found that stakeholders perceived the Kiri-Spoon to be more effective: they felt that the quantity of food acquired by the Kiri-Spoon --- and the security with which the Kiri-Spoon held that food --- made it more reliable.

Overall, half of the participants said that they preferred eating from the Kiri-Spoon while the other half preferred the regular spoon. 
Participants who liked using the Kiri-Spoon said that it was ``more secure when it came to picking up food'' and that ``the spoon was not better than the Kiri-Spoon in any way.'' 
Participants who preferred the regular spoon did not provide any specific reason for their choice and stated that there was ``nothing worse about the Kiri-Spoon.''

\p{Summary}
By including stakeholders in the design process we were able to arrive at a Kiri-Spoon that participants found as comfortable as a traditional spoon.
Consistent with our results from Section~\ref{sec:auto_experiment}, stakeholders also perceived Kiri-Spoon to be more effective than a standard utensil --- and this stakeholder viewpoint is critical as we develop assistive systems for real-world users.
However, we recognize that our results are necessarily limited because of the $N=4$ sample size.
As such, in the next section we conduct a follow-up study on users without disabilities to more precisely test the benefits of Kiri-Spoon. We wish to evaluate the comfort and effectiveness of Kiri-Spoon in both scooping and picking tasks with a larger number of participants.
\begin{figure*}[t]
    \begin{center}
        \includegraphics[width=1\textwidth]{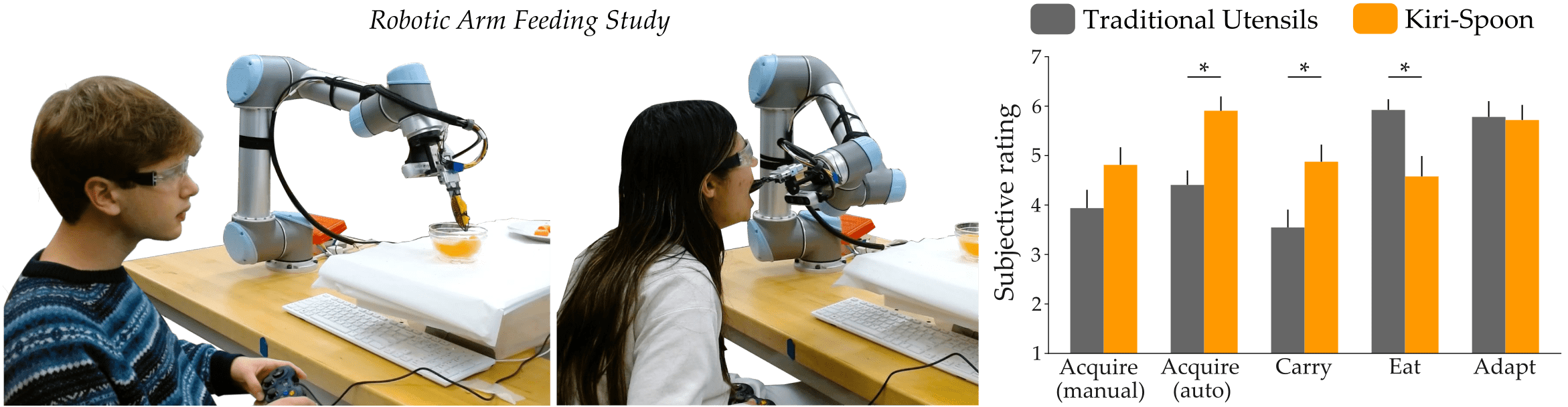}
        \vspace{-3 ex}
        \caption{Experimental setup for our comparison of mechanical and algorithmic intelligence in Section~\ref{sec:finalstudy}. We varied the robot's control algorithm and the feeding utensil, and explored the effects of both variables. (Left) Users teleoperating the robot arm to scoop food from the bowl and then eating that food from the feeding utensil. (Right) Subjective results. While users found it equally difficult to manually acquire food using traditional utensils and Kiri-Spoon, they perceived Kiri-Spoon to be significantly more effective than the traditional utensils when acquiring food autonomously. Users also rated the Kiri-Spoon to be more secure and easier to control when carrying the acquired food to their mouths. On the contrary --- perhaps because of their familiarity with spoons and forks --- users found it more comfortable to eat morsels from traditional utensils as compared to Kiri-Spoon. The error bars indicate standard error.}
        \label{fig:study_subjective}
        \vspace{-2 ex}
    \end{center}
\end{figure*}

\section{User Study with Participants without Disabilities} \label{sec:finalstudy}

To more precisely analyze the role of Kiri-Spoon within robot-assisted feeding, we conducted a final user study on participants without disabilities.
A variety of related works have developed \textit{algorithmic intelligence} that enables robot arms to dexterously manipulate utensils and feed users.
By contrast, in this paper we propose to leverage \textit{mechanical intelligence} to make assistive eating easier.
For this final study we therefore explore the independent and complementary effects of both directions.
Participants interact with a $6$-DoF UR5 \citep{ur5} robot arm to grasp, carry, and eat diverse foods.
We vary the \textit{algorithm} that controls the robot arm: in one condition users teleoperate the robot throughout the task, and in the other condition the robot leverages state-of-the-art methods to autonomously acquire and transfer foods.
We also vary the \textit{utensil} that the robot manipulates: similar to Section~\ref{sec:auto_experiment}, we compare Kiri-Spoon to traditional forks and spoons.
Overall, the goal of this section is to measure how algorithmic and mechanical advances \textit{separately} contribute to robot-assisted feeding, as well as how the \textit{combination} of control software and utensil hardware improves system performance.

\subsection{Independent Variables}
We varied the robot along two axes: (a) its \textit{feeding utensil} and (b) its \textit{control algorithm}.
For feeding utensils, we compared \textbf{traditional utensils} to \textbf{Kiri-Spoon}.
When the desired food required stabbing (i.e., grapes), the robot used the fork, and when the desired food needed scooping (i.e., cereal), the robot used the spoon.
By contrast, in the Kiri-Spoon condition the robot applied our utensil across all foods.
For the control algorithm, we compared manual teleoperation (\textbf{manual}) to autonomous food acquisition (\textbf{auto}).
During manual teleoperation participants used a joystick to modulate the position and orientation of the end-effector when acquiring the food.
On the other hand, in the autonomous mode the robot used SPANet \citep{feng2019robot, gordon2023towards} to detect the food item and choose the appropriate acquisition strategy.
After acquiring the food, participants teleoperated the robot to bring the food to their mouths. The robot maintained a speed of $0.15$ meters/second during teleoperation across all conditions.

\subsection{Participants}
We recruited $16$ adults without mobility limitations ($4$ female, $3$ non-binary, ages $22.5 \pm 2.5$ years) from the Virginia Tech community. 
All participants provided informed written consent as per the university guidelines (IRB $\# 22$-$308$).
The study followed a within-subjects design: each participant used both control algorithms (\textbf{manual} and \textbf{auto}) and interacted with both feeding utensils (\textbf{traditional} and \textbf{Kiri-Spoon}).
We balanced the order of control algorithms and feeding utensils using a Latin square design to account for ordering effects.
Put another way, the same number of participants started with \textbf{manual}, \textbf{traditional} as started with \textbf{auto}, \textbf{Kiri-Spoon}.

\subsection{Study Procedure}
Images of our experimental setup are shown in \fig{study_subjective}.
Participants interacted with a $6$-DoF UR5 robot arm \citep{ur5} that had utensils attached to its end-effector.
Using this assistive arm, participants fed themselves four different bite-sized foods.
The selected foods were a subset of the items tested in our previous experiments, but were tailored to be more appetizing for the participants.
Specifically, the foods included grapes, orange slices, cereal and canned oranges.
We selected these foods in order to test morsels with different shapes, sizes, and textures.
To acquire the grapes and orange slices the robot needed to use picking motions (i.e., leveraging a plate and fork).
Alternatively, to acquire the cereal and canned oranges --- which included the juice from the can --- the robot had to perform scooping motions (i.e., leveraging a bowl and spoon). 
Once the food had been picked or scooped the participants brought it to their mouths and ate the morsels.

After the participants were introduced to the system they had five minutes to practice manipulating the joystick and utensils.
The participants then attempted to eat each of the foods under every combination of control strategy and feeding utensil.
For a given trial, participants were allowed a maximum of three attempts.
If the user was unable to acquire, carry, and transfer the desired morsel to the mouth across all three attempts, that trial was marked as a failure and the user moved on to the next trial.

\begin{figure*}[t]
    \begin{center}
        \includegraphics[width=1\textwidth]{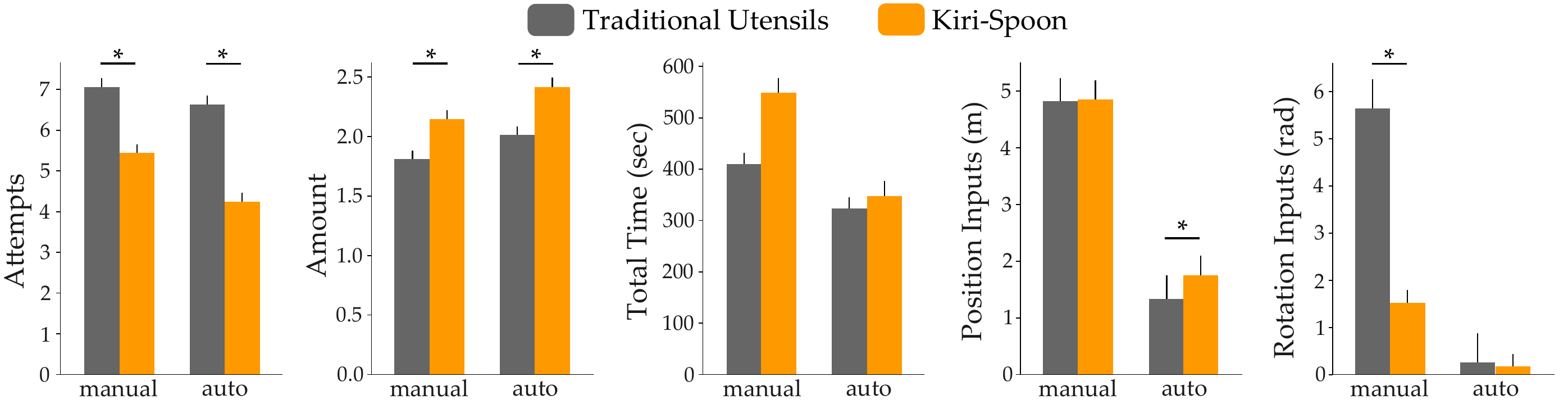}
        \vspace{-3 ex}
        \caption{Objective results from our study in Section~\ref{sec:finalstudy}. Participants interacted with a robot arm using two control algorithms: either manual teleoperation or autonomous acquisition. For each control option, we tested robots equipped with traditional utensils (i.e., forks and spoons) or Kiri-Spoon. Our results show that Kiri-Spoon reduces the number of attempts required to pick up foods, and increases the amount of food acquired. Importantly, this trend is consistent regardless of the control algorithm, suggesting that Kiri-Spoon offers a fundamental mechanical advantage. Kiri-Spoon increased the amount of time required to eat in the manual condition, likely because users were unsure how to teleoperate Kiri-Spoon. Finally, Kiri-Spoon does not require precise orientation to acquire foods, leading to fewer rotation inputs during feeding task. The error bars indicate standard error.}
        \label{fig:study_results}
        \vspace{-3 ex}
    \end{center}
\end{figure*}

\begin{table}[t!]
\centering
\caption{Survey for participants without physical disabilities (Likert scales with $7$-options)}\label{tab:questions2}
\resizebox{\linewidth}{!}{%
\begin{tabular}{l}
\hline\\[-1.5 ex]

\textbf{Acquire (Manual):}\\
Q1. It was effective and easy to pick up food manually.\\
Q2. I struggled to pick up food when controlling the robot manually.
\vspace{1 ex}\\ 

\textbf{Acquire (Autonomous):}\\
Q3. It was effective and easy to pick up food autonomously.\\
Q4. I struggled to pick up food when controlling the robot autonomously.    
\vspace{1 ex}\\ 

\textbf{Carry:}\\
Q5. It was easy to carry food using the utensil with minimal spills.\\ 
Q6. It was difficult to carry food with the utensil without dropping it.\\
Q7. I was not worried about dropping food when bringing it to my mouth.\\
Q8. I had to be precise to not drop the food when bringing it to my mouth.
\vspace{1 ex}\\ 

\textbf{Eat:}\\
Q9. It was easy to get the food off of the utensil and into my mouth.\\ 
Q10. It was difficult to get the food off of the utensil to feed myself.\\
Q11. This utensil was comfortable to eat off of.\\
Q12. This utensil was jarring to eat off of and not comfortable.
\vspace{1 ex}\\ 

\textbf{Adapt:}\\
Q13. I was able to understand and adapt to the utensil with little effort.\\ 
Q14. I was not sure how to use the utensil throughout the trial.
\vspace{0.75 ex}\\ 

\hline
\end{tabular}%
}
\vspace{-1 ex}
\end{table}

\subsection{Dependent Variables}
We recorded both objective and subjective measures to assess the independent and combined effects of control algorithm and feeding utensil.

\p{Objective Metrics} For each condition we found the number of \textit{attempts} required to successfully acquire food, the \textit{amount} of food acquired after all attempts, and the \textit{total time} required to eat all four foods using each feeding utensil and acquisition strategy. 
While the spoon and Kiri-Spoon had similar volumes, every type of food had a different size and weight. To account for this variation, we measured \textit{amount} by counting the number of items acquired in each attempt --- e.g., the number of orange slices or cereal particles --- and normalizing the counts across food types.
We also measured the complexity of operating each feeding utensil by recording the total \textit{position} and \textit{rotation} joystick inputs provided by users.

\p{Subjective Metrics} After participants completed each individual condition we administered a $7$-point Likert scale survey \citep{schrum2020four}.
The items from this survey are listed in Table~\ref{tab:questions2}.
Overall, the survey is divided into five scales: 
how easy it was to \textit{acquire} the food manually versus autonomously, how easy it was to \textit{carry} the food securely to the human's mouth, how comfortable it was to \textit{eat} the food from the utensil, and whether users were able to \textit{adapt} to that utensil.
Additionally, after users completed the experiment we asked them to indicate their preferred utensil and control algorithm using a forced-choice paradigm.

\subsection{Objective Results}
Figure~\ref{fig:study_results} summarizes our objective results. We performed a two-way ANOVA with repeated measures to evaluate the effects of \textit{feeding utensil} and \textit{control algorithm} on each of the dependent variables. 

First, we observed that users required the least number of \textit{attempts} and acquired the largest \textit{amount} of food when using Kiri-Spoon with autonomous acquisition. 
Our results indicated significant main effects for \textbf{feeding utensil} ($p < 0.001$) and \textbf{control algorithm} ($p < 0.05$) on the number of \textit{attempts}.
We also found significant main effects for \textbf{feeding utensil} ($p < 0.01$) and marginal significance for \textbf{control algorithm} ($p = 0.057$) on the \textit{amount} of food acquired.
There was no significant interaction between the two independent variables for either attempts or amount. 
Interestingly, on average, \textit{users acquired more food and needed fewer attempts when controlling the Kiri-Spoon manually as compared to using traditional utensils with autonomous control}.
This suggests Kiri-Spoon can contribute to assistive feeding independent of the control algorithm used by the robot arm.
Overall, we find that both the mechanical advantages of Kiri-Spoon and the algorithmic advancements of recent autonomous approaches enhance efficiency and success of food acquisition.

Next, we observed that while users required fewer attempts with Kiri-Spoon, they took more \textit{total time} using this utensil.
Our results indicated no significant main effect for \textbf{feeding utensil} ($p = 0.305$) but a significant main effect for \textbf{control algorithm} ($p < 0.05$) on the \textit{total time}, with a marginally significant interaction between these two independent variables ($p = 0.059$).
Put another way, we found that the time efficiency of Kiri-Spoon was dependent on the choice of acquisition strategy.
When participants had to teleoperate the robot arm, Kiri-Spoon added additional time; we suspect that this may have occurred because users were unfamiliar with how to manipulate Kiri-Spoon.
By contrast, when Kiri-Spoon was controlled autonomously, the total time was comparable to traditional utensils.

Lastly, we investigated the number of \textit{joystick inputs} that users needed to control the \textit{position} and \textit{orientation} of the feeding utensils.
As a reminder, in the manual condition participants used a joystick to teleoperate the robot arm throughout the entire task.
In the autonomous condition the robot acquired foods without any assistance --- but users still needed to leverage the joystick at the end of the task to safely move the robot closer to their mouth.
Hence, we would expect that autonomous acquisition leads to fewer inputs than manual teleoperation; and indeed, \textbf{control algorithm} had a significant main effect for both position ($p < 0.001$) and rotation ($p < 0.001$).
When comparing Kiri-Spoon to traditional utensils, users applied roughly the same number of inputs for position ($p=0.138$).
On the other hand, \textbf{feeding utensil} had a significant effect on the number of orientation inputs ($p < 0.01$).
In Section~\ref{sec:auto_experiment} we found that Kiri-Spoon could leverage its flexible structure to acquire foods without changing orientation; this trend continued here, where participants picked and scooped foods without having to tune the angle of Kiri-Spoon.

\p{Summary} Our objective results demonstrate the complementary benefits of using the mechanically superior Kiri-Spoon design with recent autonomous approaches for food acquisition. 
When controlled autonomously, Kiri-Spoon was able to acquire the most amount of food in the least number of attempts. 
It also needed fewer rotational inputs than standard utensils when being controlled manually. 
However, the additional step of having to actuate the kirigami sheet negatively impacted its time efficiency, especially during user teleoperation.
We suggest that this effect may have been caused by the novelty of our system, and perhaps users more familiar with Kiri-Spoon will not require added time.

\subsection{Subjective Results}
Given our objective results, we now change gears to focus on how users subjectively perceived the assistive robot.
Our overall subjective results are displayed in \fig{study_subjective}, Right.

Participants thought it was easier to acquire and carry foods using Kiri-Spoon as compared to traditional utensils. 
A paired t-test indicated a significant difference between traditional utensils and Kiri-Spoon when acquiring the food autonomously ($p<0.01$), and for carrying that food without spilling ($p<0.05$). 
However, users provided a higher rating ($p<0.05$) for the comfort of traditional utensils.
This is not unexpected; forks and spoons have been optimized for human comfort over thousands of years \citep{foote1934spoons}.

Despite the lack of comfort, $10$ out of $16$ users stated that they would prefer to use Kiri-Spoon over traditional utensils.
In addition, $2$ of the remaining $6$ users felt that both utensils were equal.
Participants explained their preference by stating that ``the Kiri-Spoon was much more versatile than the traditional utensils, and excelled at picking up every type of food'' even though it was ``bit less comfortable.''
One of the users also mentioned that they ``enjoyed the flexibility of the Kiri-Spoon'' which ``made picking up food easy and intuitive'', whereas the ``traditional utensils were difficult to control due to their rigid nature.''
Users who preferred the traditional utensils said that they ``enjoyed the traditional utensil because it was stagnant in their mouth.''
These comments indicate that --- at least for some users --- the mechanical advantages offered by the Kiri-Spoon outweigh its lack of comfort as compared to traditional utensils.
Lastly, when asked about their preferred control approach, $10$ out of $16$ users chose autonomous acquisition, stating that ``autonomous feeding saves more effort.'' 

\p{Future Improvements}
Some users also provided comments that present directions for improving the Kiri-Spoon in future work. 
For instance, one user stated that they wished to have ``more control options over the Kiri-Spoon'' such as the ability to ``flip it upside down to press it down on the food and close it'' for all tasks.
Another user stated that ``the Kiri-Spoon with lots of practice would be ideal in combination with a fork.''
Based on these comments, in future work, we intend to equip Kiri-Spoon with soft fork-like tines that can skewer flat foods, and then evaluate user perceptions of Kiri-Spoon after long-term use.
We hypothesize that once people get familiar with using the Kiri-Spoon --- as they are with traditional utensils --- they could feel more comfortable eating from a deformable feeding utensil.

\section{Conclusion} \label{sec:conclusion}

Assistive robot arms have the potential to improve the lives of adults with mobility limitations.
To achieve this potential, we believe that both \textit{algorithmic} and \textit{mechanical} intelligence are necessary.
In this work we therefore collaborated with stakeholders to develop a mechanical utensil specifically for robot-assisted feeding.
Our resulting mechanism (which we named Kiri-Spoon) consists of a soft kirigami sheet and a compact 1-DoF linear actuator.
At equilibrium, the kirigami sheet is a spoon-like $2$D ellipse; but when we extend this sheet, the structure deforms into a $3$D bowl that encapsulates food items.
We highlight that our design has several attractive features for adoption --- the key materials are food safe, inexpensive to manufacture, interchangeable, and washable.
Moreover, Kiri-Spoon can be deployed as both a spoon (scooping foods like cereal or soup) and as a fork (pinching foods like carrots or oranges).

To analyze Kiri-Spoon, we developed a mechanics model that relates the amount of applied force to the geometry of the kirigami sheet.
This model is challenging because the kirigami sheet consists of multiple deformable ribbons --- a boundary ribbon, discrete ribbons, and mesh ribbons.
Our theoretical model integrated each of these ribbons to provide a lower bound on the combined interaction.
Designers can leverage this model to select the correct material, geometry, and actuation for their own Kiri-Spoon.

Next, we conducted three separate experiments to evaluate how Kiri-Spoon advances robot-assisted feeding.
(1) We first compared Kiri-Spoon to traditional forks and spoons in a fully autonomous setting: here the robot arm used a state-of-the-art assistive feeding algorithm to control each utensil and acquire diverse foods.
We found that Kiri-Spoon led to higher acquisition rates for small and soft items (e.g., peas, tofu, jello) and similar acquisition rates for larger morsels (e.g., carrots, mac \& cheese).
(2) In our second experiment we attached Kiri-Spoon to a commercial assistive eating device, and brought the resulting system to a local center for adults with physical disabilities.
A caregiver and $N=4$ participants compared the device with and without Kiri-Spoon; their results suggest that users perceive Kiri-Spoon to be about as comfortable as a traditional spoon, but more effective at acquiring and transferring foods.
(3) To better analyze these results we conducted a follow-up study on users without physical disabilities.
For this final study we attached the Kiri-Spoon to a UR5 robot arm, and varied two separate factors: the algorithm the robot used to control the utensil, and the utensil the robot was equipped with.
Our results suggest that both algorithm and utensil have an impact on performance.
Interestingly, the improvements caused by Kiri-Spoon in efficiently acquiring the foods were larger than the improvements caused by using an autonomous feeding algorithm --- indicating the importance of mechanical intelligence.
Overall, the combination of both state-of-the-art algorithms and our Kiri-Spoon utensil led to the most effective robot-assisted feeding.

\p{Limitations} Taken together, our theoretical and experimental results suggest that Kiri-Spoon can meaningfully advance robot-assisted feeding by making the process of acquiring, carrying, and transferring foods more robust.
However, we also identified some areas for improvement.
Specifically, we found that Kiri-Spoon sometimes failed to grasp foods that had a large, flat geometry (e.g., lettuce).
Kiri-Spoon fell short here because it was unable to scoop or pinch the lettuce without the food slipping away --- whereas a traditional fork could just skewer this morsel.
Inspired by sporks, in our future work we will explore adding soft fork-like tines to Kiri-Spoon to better handle these edge cases.
We are also interested in using Kiri-Spoon on more complex plates with multiple different morsels.
If the food types are separated on the same plate, then our current approach is sufficient --- for instance, in Section~\ref{sec:userstudy} participants ate from multiple bowls each with different items.
But future work should focus on settings where these foods are mixed together --- i.e., noodles with meatballs --- and the utensil needs to handle this combination.

%%%%%%%%%%%%%%%%%%%%%%%%%%%%%%%%%%%%%%%%%%%%%%%%%%%%%%%%%%%%%%%%%%%%%%%%%%%%%%%%%%%%%%%%

\begin{dci}
The author(s) declared no potential conflicts of interest with respect to the research, authorship, and/or publication of this article.
\end{dci}

%%%%%%%%%%%%%%%%%%%%%%%%%%%%%%%%%%%%%%%%%%%%%%%%%%%%%%%%%%%%%%%%%%%%%%%%%%%%%%%%%%%%%%%%

\begin{funding}
This work is supported in part by NSF Grants $\#2205241$ and $\#2337884$. This paper is dedicated to Tyler Schrenk.
\end{funding}

%%%%%%%%%%%%%%%%%%%%%%%%%%%%%%%%%%%%%%%%%%%%%%%%%%%%%%%%%%%%%%%%%%%%%%%%%%%%%%%%%%%%%%%%

\begin{sm}
\p{Experimental Videos} Videos of our system and experiments are publicly available at: \url{https://youtu.be/ZJpQREdTz80} 

\smallskip

\p{Design Files} The CAD models for our Kiri-Spoon and its kirigami sheet are publicly available at: \url{https://github.com/VT-Collab/Kiri-Spoon}
\end{sm}

%%%%%%%%%%%%%%%%%%%%%%%%%%%%%%%%%%%%%%%%%%%%%%%%%%%%%%%%%%%%%%%%%%%%%%%%%%%%%%%%%%%%%%%%

\bibliographystyle{SageH}
\bibliography{references.bib}
\appendix

\section{Appendix}

\subsection{Simulation of Boundary Deformation}\label{sec:bend}

In Section~\ref{sec:model} we propose a theoretical model for tensile force required to actuate the kirigami structure of Kiri-Spoon. 
One component of this tensile force is the force required to bend the \textit{boundary} of the kirigami sheet. 
We compute this force, $F_{bend}$, using \eq{f_bend} based on the bending theory of circular rings.
In order to apply this theory we assume that the radius of curvature of the boundary ring is equal to the initial radius $r$ of the kirigami sheet at its equilibrium.

As the boundary deforms into an ellipse, however, its radius of curvature changes. 
In particular, the radius of curvature decreases at the ends of the major axis along which we apply the tensile force. 
Based on \eq{f_bend} we know that the $F_{bend}$ is inversely proportional to the radius of curvature. Therefore, the actual force required to bend the elliptical boundary should be higher than the force computed using the initial radius of the circular kirigami sheet $r$.
In other words, our theoretic model provides a lower bound on the actual bending force. 
Here we present physics simulation results to validate this claim.

\begin{figure}[h!]
    \centering
    \includegraphics[width=1\linewidth]{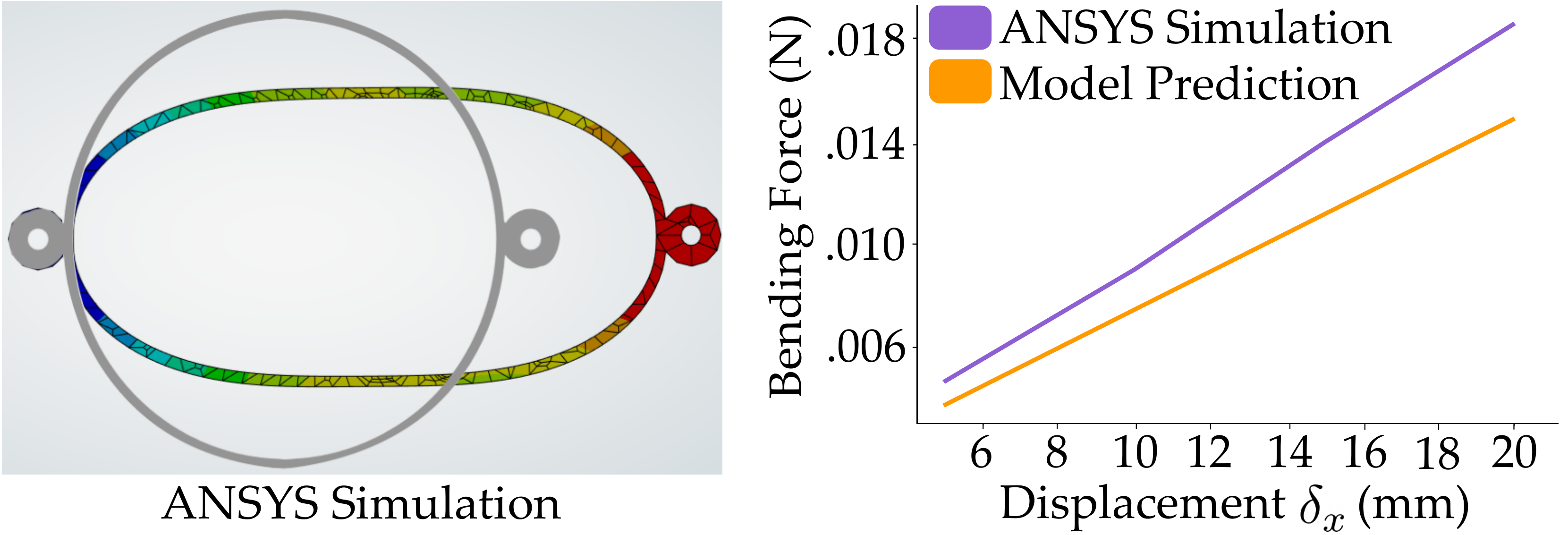}
    \caption{Physics simulation in support of \eq{f_bend}. (Left) ANSYS simulation environment. The initial circular boundary is shown in grey, while the colored ellipse depicts the deformed elliptical boundary for a displacement of 20 millimeters. (Right) Simulation results showing that the force predicted by our model in \eq{f_bend} is a lower bound on the actual tensile force required to bend the boundary ribbon.}
    \label{fig:ansys}
\end{figure}

We simulate the bending of a circular ring in ANSYS Mechanical \citep{ansys}, a finite element analysis software for structural deformations.
A visualization of this simulation is shown in \fig{ansys}.
For our testing we use a 3D model of a ring with the same material properties and dimensions as kirigami sheet A in Table~\ref{tab:sheet}. 
We define one end of the circular ring as a fixed support, and apply an incremental tensile load on the opposite end. 
As the ring deforms, we measure the displacement along the major axis. 
We then compare the force applied in simulation to the force computed by our model using \eq{f_bend}. 
Our results are shown in Figure~\ref{fig:ansys}, Right.
From this test we find that the force required to deform the boundary in simulation is higher than the force estimated by our model; this result aligns with our claim that $F_{bend}$ is a \textit{lower bound} on the bending force.

\subsection{Derivation for Discrete Ribbons Bending}\label{sec:discrete_component}

When the kirigami sheet is actuated, the boundary bends and pushes on the enclosed discrete ribbons. 
These discrete ribbons oppose the deformation of the boundary, and so we must apply an additional tensile force $F_{discrete}$ to overcome their resistance.
In Section~\ref{sec:discrete} we outlined our derivation for $F_{discrete}$. 
Here we provide additional details and analysis for computing this tensile force.

\begin{figure}[h]
    \centering
    \includegraphics[width=1\linewidth]{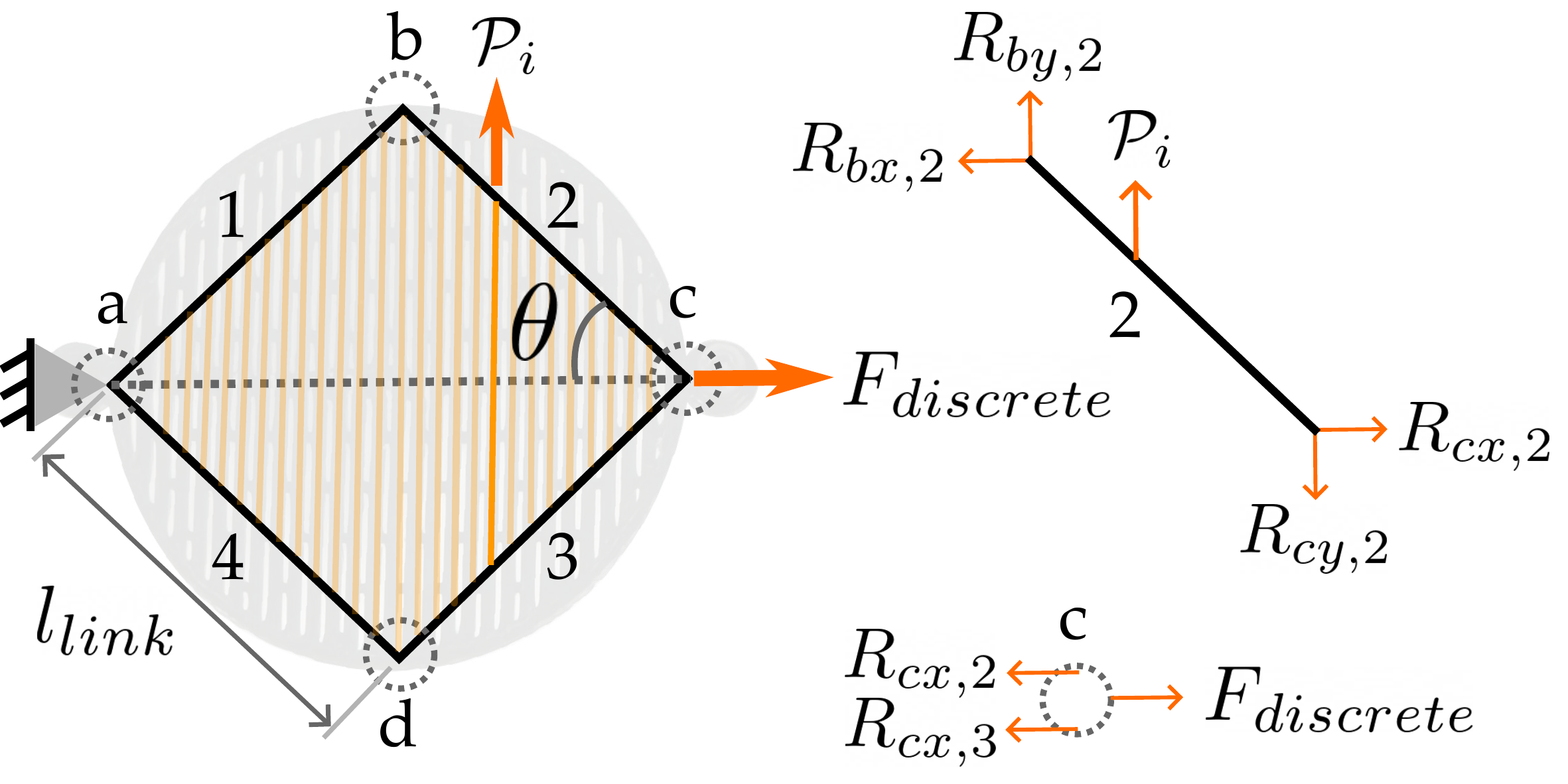}
    \caption{Finding the resistance force caused by the discrete ribbons. (Left) Four-bar linkage model. Joints $a$ and $c$ specify the ends of the major axis of the elliptical boundary layer, while joints $b$ and $d$ specify its minor axis. The joints are connected by rigid links $1$, $2$, $3$ and $4$. When joint $c$ is pulled along the major axis by a tensile force $F_{discrete}$, the links bring joints $b$ and $c$ closer, reducing the width of the boundary along the minor axis. This deforms the discrete ribbons which exert an opposing force $\mathcal{P}$ on the links. (Right) Free body diagrams of link $2$ and joint $c$.}
    \label{fig:four_bar}
\end{figure}

We will first compute the tensile force needed to overcome a single discrete ribbon. 
Consider a kirigami sheet with $n_{d}$ discrete ribbons and let $\mathcal{P}_{i}$ be the opposing force exerted by the $i$-th discrete ribbon. 
We start by modeling the boundary as a four-bar linkage where joints $a$ and $c$ align with the ends of the major axis of the elliptical boundary, and joints $b$ and $d$ mark its minor axis (see \fig{four_bar}).
As such, half of the discrete ribbons are to the left of joints $b$ and $d$, and the other half are to their right. The joints are sequentially connected by rigid links of length $l_{link}$. 
As shown in Figure~\ref{fig:four_bar}, $F_{discrete}$ is applied at joint $c$ by keeping the position of joint $a$ fixed, while the opposing forces from the discrete ribbons act on the links between these joints. 

Let the $i$-th discrete ribbon apply a force $\mathcal{P}_{i}$ on link $2$. This force results in a clockwise moment $M_{i,2}$ at joint $c$:
\begin{equation}
    M_{i,2} = \mathcal{P}_{i} \cdot l_{link} \bigg(\frac{i}{\lfloor n_{d}/2 \rfloor + 1}\bigg)\cos\theta \label{eq:discrete_single}
\end{equation}
Here $l_{link} \left(i / \lfloor n_{d}/2 \rfloor + 1\right)$ is the distance between joint $c$ and the point at which the force $\mathcal{P}_{i}$ acts on the link, and $\theta$ is the angle formed by the link with the tensile force direction.

\eq{discrete_single} calculates the moment due to a single discrete ribbon on link $2$.
To obtain the moment due to all discrete ribbons that act on link $2$ we can compute their sum:
\begin{equation}
    M_{2} = \sum_{i=1}^{ \lceil n_{d}/2 \rceil} \mathcal{P}_{i} \cdot l_{link} \left(\frac{i}{\lfloor n_{d}/2 \rfloor + 1}\right) \cos\theta \label{eq:discrete_all}
\end{equation}

Next, we need to establish how this moment relates to the tensile force $F_{discrete}$. 
The combined moment $M_{2}$ is counteracted by the moment due to the reaction forces $R_{bx, 2}$ and $R_{by, 2}$ at the ends of the link. 
From \fig{four_bar} we see that $R_{bx, 2} = R_{cx, 2}$, while $R_{by, 2}$ can be computed by balancing the forces across all links. 
In this case $R_{by, 2} = 0$. 
Therefore, the moment due to all discrete ribbons on link $2$ is equal and opposite to the moment due to $R_{bx, 2}$, yielding:
\begin{equation}
    M_{2} = R_{bx, 2} \cdot l \sin\theta = R_{cx, 2} \cdot l \sin\theta \label{eq:discrete_moment}
\end{equation}
We can now connect $R_{cx, 2}$ to $F_{discrete}$ by balancing the forces at joint $c$. 
Joint $c$ is connected to two links, $2$ and $3$.
Therefore, the tensile force applied at joint $c$ is equal to the sum of the reaction forces due to both the links: $F_{discrete} = R_{cx, 2} + R_{cx, 3}$. 
But because our four-bar linkage model is symmetric, $R_{cx, 2} = R_{cx, 3}$ and so:
\begin{equation}
F_{discrete} = 2R_{cx, 2} \label{eq:discrete_joint}
\end{equation}

In summary, by combining the Equations~(\ref{eq:discrete_all}),~(\ref{eq:discrete_moment}), and~(\ref{eq:discrete_joint}), we obtain the additional tensile force due to the discrete ribbons.
This result is \eq{f_discrete} in Section~\ref{sec:discrete}.

%%%%%%%%%%%%%%%%%%%%%%%%%%%%%%%%%%%%%%%%%%%%%%%%%%%%%%%%%%%%%%%%%%%%%%%%%%%%%%%%%%%%%%%%
\end{document}